\documentclass{article}

\usepackage{microtype}
\usepackage{graphicx}
\usepackage{subfigure}
\usepackage{booktabs} 

\usepackage{hyperref}

\usepackage[accepted]{icml2025}

\usepackage{amsmath}
\usepackage{amssymb}
\usepackage{mathtools}
\usepackage{amsthm}

\usepackage[capitalize,noabbrev]{cleveref}

\theoremstyle{plain}

\theoremstyle{definition}

\theoremstyle{remark}

\usepackage[textsize=tiny]{todonotes}
\usepackage{CJKutf8}

\usepackage[utf8]{inputenc} 
\usepackage[T1]{fontenc}    
\usepackage{url}            
\usepackage{nicefrac}       
\usepackage{lipsum}		
\usepackage{doi}

\usepackage{xcolor}         

\usepackage{colortbl}
\usepackage{times}
\usepackage{microtype} %
\usepackage{balance} 
\usepackage{algorithm}
\usepackage{algorithmic}

\usepackage{hyperref}

\usepackage{helvet}
\usepackage{courier}

\usepackage{makecell}
\usepackage{graphicx}
\usepackage{color}
\usepackage{amsfonts}
\usepackage{amsmath}
\usepackage{amssymb}

\usepackage{multirow}
\usepackage{booktabs}
\usepackage{wrapfig}

\DeclareMathAlphabet\mathbfcal{OMS}{cmsy}{b}{n}

\def\0{{\bf 0}}
\usepackage{ dsfont }
\def\1{\mathds{1}}

\def\bw{{\bf w}}

\def\exp{{\mathrm{exp}}}

\def\bw{{\bf w}}

\usepackage{enumitem}

\usepackage{array}
\usepackage{tabu}
\makeatletter
\newcommand{\nipstophline}{%
	\noalign {\ifnum 0=`}\fi \hrule height 4pt
	\futurelet \reserved@a \@xhline
}
\newcommand{\nipsbottomhline}{%
	\noalign {\ifnum 0=`}\fi \hrule height 1pt
	\futurelet \reserved@a \@xhline
}

\def\xie{\textcolor{black}}

\def\TODO{\textcolor{black}}

\def\green{\textcolor{green}}
\def\red{\textcolor{red}}

\def\bw{\textcolor{black}}

\definecolor{mygray}{RGB}{200, 200, 200}

\usepackage{threeparttable}

\icmltitlerunning{FG-CLIP: Fine-Grained Visual and Textual Alignment}

\begin{document}

\twocolumn[
\icmltitle{FG-CLIP: Fine-Grained Visual and Textual Alignment}



\icmlsetsymbol{equal}{*}

\begin{icmlauthorlist}
\icmlauthor{Chunyu Xie}{yyy,equal}
\icmlauthor{Bin Wang}{yyy,equal}
\icmlauthor{Fanjing Kong}{yyy}
\icmlauthor{Jincheng Li}{yyy}
\icmlauthor{Dawei Liang}{yyy}
\icmlauthor{Gengshen Zhang}{yyy}

\icmlauthor{Dawei Leng}{yyy}
\icmlauthor{Yuhui Yin}{yyy}
\end{icmlauthorlist}

\begin{center} 
Code: https://github.com/360CVGroup/FG-CLIP\\ 
Model: https://huggingface.co/qihoo360/fg-clip-large\\
Dataset: https://huggingface.co/datasets/qihoo360/FineHARD
\end{center}

\icmlaffiliation{yyy}{360 AI Research}

\icmlcorrespondingauthor{Dawei Leng}{lengdawei@360.cn}

\icmlkeywords{Machine Learning, ICML}

\vskip 0.3in
]



\printAffiliationsAndNotice{\icmlEqualContribution} 

\begin{abstract}
Contrastive Language-Image Pre-training (CLIP) excels in multimodal tasks such as image-text retrieval and zero-shot classification but struggles with fine-grained understanding due to its focus on coarse-grained short captions. To address this, we propose Fine-Grained CLIP (FG-CLIP), which enhances fine-grained understanding through three key innovations. First, we leverage large multimodal models to generate 1.6 billion long caption-image pairs for capturing global-level semantic details. Second, a high-quality dataset is constructed with 12 million images and 40 million region-specific bounding boxes aligned with detailed captions to ensure precise, context-rich representations. Third, 10 million hard fine-grained negative samples are incorporated to improve the model's ability to distinguish subtle semantic differences. We construct a comprehensive dataset, termed FineHARD, by integrating high-quality region-specific annotations with hard fine-grained negative samples. Corresponding training methods are meticulously designed for these data. Extensive experiments demonstrate that FG-CLIP outperforms the original CLIP and other state-of-the-art methods across various downstream tasks, including fine-grained understanding, open-vocabulary object detection, image-text retrieval, and general multimodal benchmarks. These results highlight FG-CLIP's effectiveness in capturing fine-grained image details and improving overall model performance. The data, code, and models are available at https://github.com/360CVGroup/FG-CLIP.
\end{abstract}    
\section{Introduction}
\label{intro}
The integration of vision and language \cite{flamingo,ramesh2022hierarchical,lin2023pmc,gabeff2023wildclip} has been a long-standing goal in artificial intelligence, aiming to develop models that can understand and reason about the world in a visually and linguistically rich manner. Recent advances in multimodal pre-training, such as CLIP~\cite{openaiclip}, have made significant strides in this direction by learning joint representations of images and text through contrastive learning. These models have achieved state-of-the-art performance in a variety of downstream tasks, including image-text retrieval \cite{pan2023prior,sun2023alphaclip,zhang2025long}, image captioning \cite{clipcap,li2023monkey,yao2024minicpm}, and visual question answering \cite{li2023blip,parelli2023clip,team2024gemini,wang2025iaa}. However, despite their impressive capabilities, these models often struggle with fine-grained details, particularly in recognizing object attributes and their relationships.

Recent works~\cite{liu2023efficient,wu2024lotlip,zhang2025long,zheng2025dreamlip,jingfineclip} point out two primary reasons for the limitations in CLIP's fine-grained learning capability. First, the original CLIP model's text encoder supports only up to 77 tokens, restricting its capacity to process detailed descriptions and hindering its ability to capture nuanced textual information. Second, CLIP aligns entire images with corresponding text descriptions, making it challenging to extract valuable region-specific representations from visual features. Consequently, the model struggles to achieve fine-grained alignment between image regions and their corresponding textual attributes, limiting its effectiveness in complex recognition scenarios.

To address these issues, researchers have proposed extending the positional encoding to support longer token sequences~\cite{wu2024lotlip,zhang2025long,zheng2025dreamlip} and integrating object detection datasets into CLIP training~\cite{zhong2022regionclip,jingfineclip}. By aligning bounding boxes with category labels, these methods aim to enhance regional feature extraction. Although these approaches have shown some improvements, they still fall short in fine-grained visual recognition and open-vocabulary object detection. Existing methods~\cite{jingfineclip,zhang2025long} typically introduce relatively few long captions, usually on a million scale, which is inadequate for effective learning of fine-grained details. Additionally, aligning image regions with category labels limits semantic diversity, restricting the model's generalization to open-world scenarios. Furthermore, the lack of hard fine-grained negative samples limits the model's ability to distinguish between objects of the same category but with different attributes.  In this work, we introduce Fine-Grained CLIP (FG-CLIP), \bw{a novel approach} designed to enhance CLIP's fine-grained understanding capabilities through three key innovations.

First, we significantly enhance global-level semantic alignment by generating long captions using state-of-the-art large multimodal models (LMMs)~\cite{hong2024cogvlm2}. This process introduces 1.6 billion long caption-image pairs, providing an unprecedented scale of data that allows FG-CLIP to capture nuanced details at the global-level semantic layer, thereby enhancing its ability to perceive complex and detailed information.

Second, to improve fine-grained alignment between images and text, we develop a high-quality visual grounding dataset. This dataset includes detailed descriptions for 40 million bounding boxes across 12 million images, ensuring that each region is precisely annotated with context-rich captions. By creating such an extensive and richly annotated dataset, we enable the model to learn precise and contextually rich representations, significantly enhancing its performance on tasks that require fine-grained understanding.

Third, to further enhance model robustness and discrimination abilities, we introduce a large-scale corpus of 10 million hard fine-grained negative samples. By incorporating these challenging negative samples into the training process, FG-CLIP learns to distinguish subtle differences in semantically similar but distinct pairs, thereby significantly improving its performance across various downstream tasks. We integrate the high-quality visual grounding data and hard fine-grained negative samples as a whole dataset called FineHARD.

Compared to previous methods, FG-CLIP demonstrates significant improvements across a wide range of benchmark tasks. Our comprehensive enhancements enable the model to achieve superior performance in capturing nuanced visual details, as evidenced by our state-of-the-art results on tasks such as fine-grained understanding, bounding box classification, long caption image-text retrieval, and open-vocabulary object detection. Moreover, when utilized as the backbone for LMMs~\cite{llava}, FG-CLIP also demonstrates performance improvements in tasks involving attribute analysis~\cite{hudson2019gqa}, object localization~\cite{refcoco}, and reducing output hallucination~\cite{pope}. \TODO{We provide visualization results in Appendix \ref{VC} to demonstrate the improvement in fine-grained understanding.} These results highlight FG-CLIP's effectiveness in capturing fine-grained image details and improving overall model performance. 
\TODO{To facilitate future research and application, we make the models, datasets, and code publicly available at https://github.com/360CVGroup/FG-CLIP.}

\begin{figure*}[!htbp]
  \centering   \includegraphics[width=1.0\linewidth]{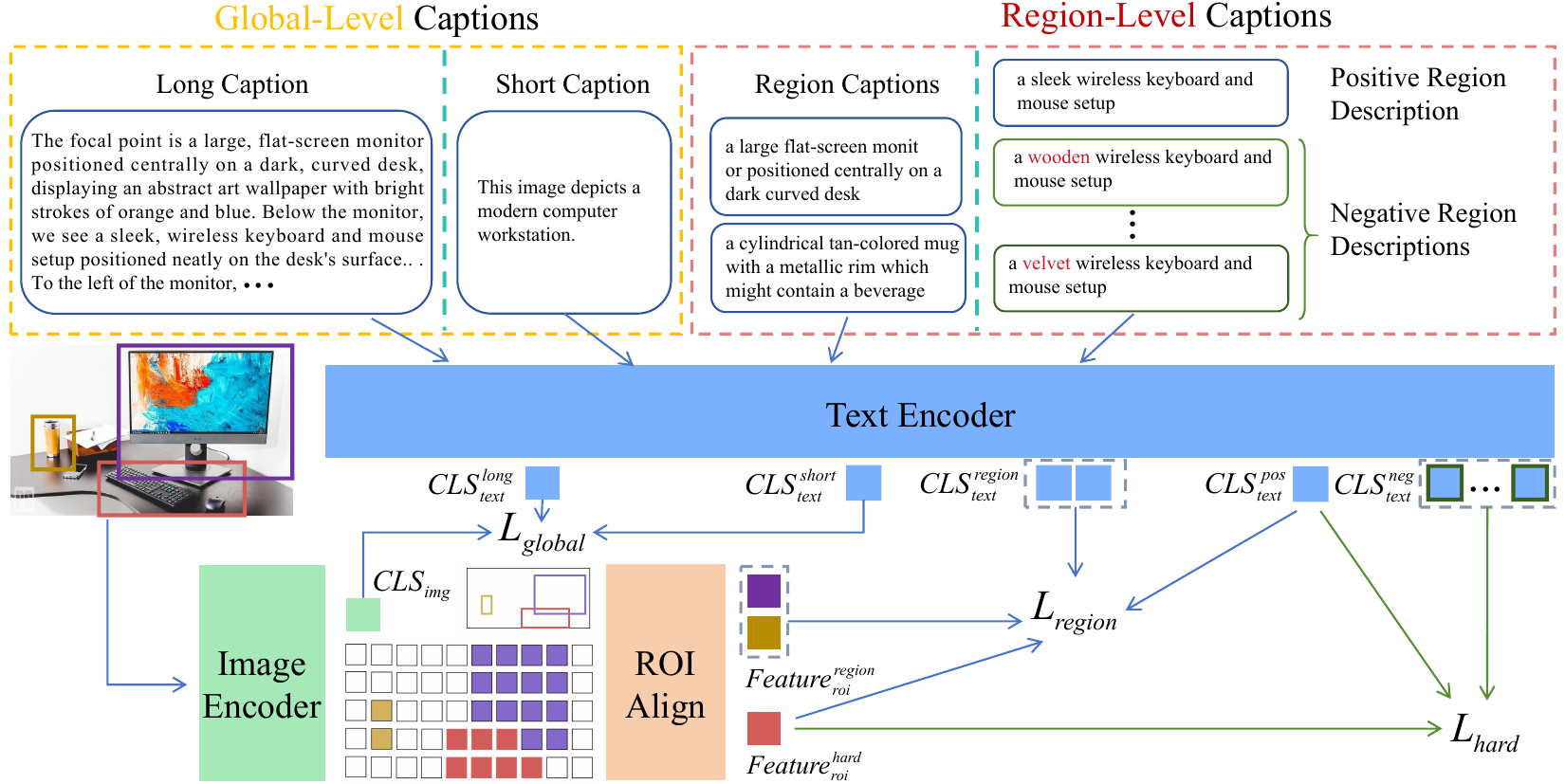}
    \caption{Overview of the FG-CLIP. $CLS_{img}$ denotes the image class features output by the Vision Transformer (ViT), while $CLS_{text}$ represents the class features summarized by the text model for multiple inputs, including long captions, short captions, region captions, and positive\&negative descriptions of specific regions within images. FG-CLIP's training proceeds in two stages: the first stage leverages global-level caption-image pairs to achieve initial fine-grained alignment, while the second stage supplements these with additional region-level captions, including detailed region captions and positive/negative region descriptions to further refine the alignment.}
    
    \label{fig:arch}
\end{figure*}

\section{Related Work}
\label{sec:related_work}

\subsection{Contrastive Language-Image Pre-training}
Contrastive learning has emerged as a powerful paradigm in multimodal pre-training, significantly advancing the field of image-text alignment. Models like CLIP have revolutionized this area by leveraging large-scale image-text pairs to learn rich representations without explicit supervision. CLIP achieves this through a dual-encoder architecture that maps images and their corresponding text descriptions into a shared embedding space, where semantically similar pairs are pulled closer together while dissimilar pairs are pushed apart. This approach not only simplifies data labeling but also enables zero-shot transfer to downstream tasks, demonstrating impressive performance on various benchmarks such as image classification~\cite{deng2009imagenet,recht2019imagenet} and image-text retrieval~\cite{young2014image,lin2014microsoft,urbanek2024picture,chen2025sharegpt4v}.

\subsection{Fine-Grained Understanding}
Despite its success, CLIP faces limitations in handling fine-grained visual details. Its text encoder is constrained to 77 tokens, limiting its capacity to process detailed and complex descriptions. Additionally, CLIP aligns entire images with corresponding text, making it challenging to extract valuable region-specific representations. To address these limitations, models like LongCLIP~\cite{zhang2025long} extend the maximum token length of the text encoder, enabling it to handle longer and more detailed textual information. GLIP~\cite{li2021glip} and RegionCLIP~\cite{zhong2022regionclip} introduce grounding data, enhancing the model's ability to align specific regions within images with corresponding text, thereby improving performance on downstream detection tasks~\cite{xie2018memory,gupta2019lvis,zhou2022detecting,minderer2024scaling}. 
However, even with these improvements, existing models still struggle to fully capture and align fine-grained features across diverse datasets.

\subsection{Image-Text Datasets}
Image-text datasets~\cite{gu2022wukong,xie2023ccmb,pmlr-v235-fu24b} play a pivotal role in the performance of multimodal models. While existing datasets such as LAION~\cite{schuhmann2021laion,schuhmann2022laion}, COCO~\cite{lin2014microsoft}, Flickr30K~\cite{young2014image}, and Conceptual Captions~\cite{sharma2018conceptual,changpinyo2021conceptual} offer valuable resources, they often emphasize general scene descriptions, neglecting fine-grained details critical for advanced applications. Researchers have adopted several strategies to mitigate these limitations. One approach involves leveraging advanced large multimodal models~\cite{idefics3,wang2024qwen2,wu2024deepseek,chen2024internvl,team2024gemini} to refine and enrich text descriptions through recaptioning. For instance, LongCLIP~\cite{zhang2025long} utilizes 1 million long caption-image pairs from ShareGPT4V~\cite{chen2025sharegpt4v}, and FineCLIP~\cite{jingfineclip} constructs a dataset of 2.5 million long caption-image pairs. Although these efforts enhance data richness, they remain limited in scale compared to the vast amount of data in the image-text field.
Another strategy is to implement pseudo-labeling pipelines using pre-trained object detection models~\cite{li2023makes,ma2024groma,hou2024salience} to automatically generate fine-grained pseudo-labels for region boxes, similar to the GRIT dataset utilized in Kosmos-2~\cite{peng2023kosmos}. These methods help improve region-specific alignment but may introduce noise due to automated labeling.

Another significant challenge is the scarcity of hard fine-grained negative samples. Existing datasets predominantly consist of positive examples that are relatively easy to distinguish, limiting the model's ability to learn subtle variations. The absence of hard negative samples impedes true fine-grained understanding, as models struggle to discern small but meaningful differences in visual and textual features. Addressing this gap is essential for developing models capable of reliably performing fine-grained recognition and alignment tasks, thereby enabling them to handle the nuanced distinctions necessary for advanced applications.

\section{Approach}
\label{Approach}

\subsection{Fine-Grained CLIP}
\label{sec:fine_grained_clip}

Figure \ref{fig:arch} provides an overview of Fine-Grained CLIP (FG-CLIP). Our proposed FG-CLIP extends the traditional dual-encoder architecture of CLIP to better capture fine-grained details in images and text. We leverage a two-stage training paradigm to achieve this enhancement. In the first stage, FG-CLIP focuses on aligning global representations of images and text using only global contrastive learning. The second stage builds on this foundation by introducing regional contrastive learning and hard fine-grained negative samples learning, leveraging region-text data to further refine the model's understanding of fine-grained details.


\paragraph{Global Contrastive Learning.}
Global contrastive learning aims to enhance the model's fine-grained understanding by introducing a method of augmenting long caption alignment utilizing Large Multimodal Models (LMMs). This approach generates additional long captions that provide richer context and finer-grained descriptions. The inclusion of long captions enables the model to perceive and align with global-level semantic details, thereby enhancing fine-grained understanding and context awareness. In addition, we retain the alignment of short caption-image pairs. The long captions complement these short captions, ensuring that the model learns from both detailed, nuanced long captions for complex semantic information and concise, direct short captions for basic concepts. This dual approach improves the model's overall performance in capturing a broader spectrum of visual information.


In our framework, both short and long captions are aligned with images by utilizing the [CLS] token features extracted from the text encoder for the captions and the [CLS] token features from the image encoder for the images. To accommodate longer and more detailed captions while preserving the alignment of short captions, position embeddings of FG-CLIP's text encoder are extended. Specifically, for sequences shorter than or equal to 20 tokens, we use the original position embedding directly. For longer sequences, we apply linear interpolation with a factor of 4 for positions beyond 20, extending the maximum length from 77 to 248 tokens. This modification ensures that the model can effectively handle longer, more descriptive text while maintaining computational efficiency. 

During each training step, the model employs both a short caption and a long caption for every image to ensure comprehensive and fine-grained understanding. Given an image-text pair, the outputs of both encoders are embeddings $v \in \mathbb{R}^d$ for images and $t \in \mathbb{R}^d$ for text, where $d$ is the dimensionality of the embedding space. We compute the similarity between each pair using the cosine similarity metric:
\begin{align}
s(v, t) = \frac{v \cdot t^T}{\|v\|\|t\|}.
\end{align}
The objective function for global contrastive learning is based on the InfoNCE loss~\cite{he2020momentum}, which maximizes the similarity between matching pairs while minimizing the similarity between mismatched pairs. Specifically, the loss for a batch of $N$ image-text pairs is given by:

\begin{align}
L_{global} = -\frac{1}{2N}\sum_{i=1}^{N} (\log \frac{\exp(s(v_i, t_i)/\tau)}{\sum_{j=1}^{N} \exp(s(v_i, t_j)/\tau)} \notag\\
+\log \frac{\exp(s(t_i, v_i)/\tau)}{\sum_{j=1}^{N} \exp(s(t_i, v_j)/\tau)}),
\end{align}

where $\tau$ is a learnable temperature parameter. This global contrastive learning significantly improving its detail perception capabilities in both granular and holistic contexts.


\paragraph{Regional Contrastive Learning.}
Regional contrastive learning focuses on aligning specific regions within images with corresponding text segments. To achieve this, we employ RoIAlign~\cite{he2017mask} to extract region-specific features from the image. These extracted features are then processed by applying average pooling over the tokens within each detected region, resulting in a set of region embeddings $\{r_k\}_{k=1}^{K}$, \TODO{where $K$ denotes the total number of valid bounding boxes across all images within a batch.}
This approach differs from global contrastive learning, which typically relies on the [CLS] token for deriving image-level features. For text, we segment the full-image caption into phrases or sentences that correspond to individual bounding boxes, obtaining text embeddings $l_k$. The regional contrastive loss is defined as:

\begin{align}
L_{regional} = -\frac{1}{2K}\sum_{i=1}^{K} (\log \frac{\exp(s(r_i, l_i)/\tau)}{\sum_{j=1}^{K} \exp(s(r_i, l_j)/\tau)} \notag\\
+\log \frac{\exp(s(l_i, r_i)/\tau)}{\sum_{j=1}^{K} \exp(s(l_i, r_j)/\tau)}).
\end{align}

This encourages the model to learn fine-grained alignments between specific regions and textual descriptions.




\paragraph{Hard Fine-Grained Negative Samples Learning.}
To address the scarcity of challenging fine-grained negative samples, we introduce a hard negative \bw{mining} strategy. We define hard negative samples as those that are semantically close but not identical to the positive sample. These hard negatives are constructed by rewriting the descriptions of bounding boxes, modifying certain attributes to create subtle differences. The specific process of obtaining hard fine-grained negative samples can be found in Section \ref{sec: curated dataset}.

To incorporate hard negative samples into the learning process, we extend the loss function to include a term for hard negatives. For each region-text pair, we compute the similarity between the regional feature and both the positive description and the corresponding negative sample descriptions. The hard negative loss $ L_{hard} $ is defined as:
\begin{equation}
L_{hard} = -\frac{1}{K}\sum_{i=1}^{K}  \log \frac{\exp(s(r_i, l_{i,1})/\tau)}{\sum_{j=1}^{M} \exp(s(r_i, l_{i,j})/\tau)},\\
\end{equation}
where $M$ denotes the total number of captions for each region, with $ j=1 $ corresponding to the positive sample, and $ j>1 $ corresponding to the negative samples.

In the second stage, we integrate all three components: Global Contrastive Learning, Regional Contrastive Learning, and Hard Fine-Grained Negative Samples Learning, to ensure comprehensive and nuanced alignment tasks. The learning objective in the second stage combines these elements:

\begin{equation}
L =  L_{global}+ \alpha * L_{regional}+ \beta *L_{hard}. 
\end{equation}

Here, the hyperparameters $\alpha$ and $\beta$ are set to 0.1 and 0.5, respectively, to balance the regional contrastive loss and the hard negative loss, ensuring that each loss operates on similar scales.

This integrated approach ensures that FG-CLIP not only captures global-level semantic details but also distinguishes subtle differences in semantically similar pairs, enhancing its overall performance across various downstream tasks.

\subsection{Curated Dataset}
\label{sec: curated dataset}
In this section, we describe the meticulous process of curating datasets for our FG-CLIP model, emphasizing both scale and quality to address the limitations of existing models in fine-grained understanding.

\paragraph{Enhancing LAION-2B Data with Detailed Recaptioning.} In the first stage of training, we utilize an enhanced version of the LAION-2B dataset~\cite{schuhmann2022laion}, where images are recaptioned with detailed descriptions generated by large multimodal models, i.e., CogVLM2-19B~\cite{hong2024cogvlm2}. This approach generates more detailed and contextually rich captions, crucial for capturing subtle differences in visual content. The original LAION-2B dataset often suffers from overly generic or imprecise captions, leading to suboptimal performance in fine-grained tasks. For instance, an image of a bird might be described as "a bird", without specifying the species or environment. Such generic captions limit the model's ability to recognize fine details.

By leveraging advanced large multimodal models, we generate detailed descriptions that not only identify objects but also provide rich contextual information about their attributes, actions, and relationships within the scene. For instance,  rather than a generic description like "a bird", our refined captions read "a red-winged blackbird perched on a tree branch in a park." Utilizing a cluster of 160$\times$910B NPUs, the data processing is completed in 30 days. An ablation study detailed in Section \ref{Sec: ablation} evaluates the impact of using these high-quality, detailed captions. The results demonstrate significant improvements in model performance across various tasks, underscoring the critical role of large-scale, high-quality text annotations in enhancing both model accuracy and context understanding.
\paragraph{Fine-Grained Visual Grounding+Recaption+Hard Negative Dataset (FineHARD).} For the second stage, we develop a high-quality visual grounding dataset named \textbf{FineHARD}, \bw{featuring precise region-specific captions and hard negative samples}. We curate the overall dataset based on GRIT~\cite{peng2023kosmos} images. The process begins with generating detailed image captions using CogVLM2-19B~\cite{hong2024cogvlm2}, ensuring comprehensive and \bw{nuanced} descriptions that capture the full context of each image. Following~\cite{peng2023kosmos}, we then use SpaCy~\cite{honnibal2020spacy} to parse the captions and extract the referring expressions. Subsequently, the images and referring expressions are fed into the pretrained object detection model, i.e., Yolo-World~\cite{cheng2024yolo} to obtain the associated bounding boxes. Non-maximum suppression is applied to eliminate overlapping bounding boxes, retaining only those with predicted confidence scores higher than 0.4. This process results in 12 million images and 40 million bounding boxes with fine-grained region captions. We provide examples of the images and their corresponding captions in Appendix \ref{TSample}.

Next, to create challenging fine-grained negative samples, we modify attributes of bounding box descriptions while keeping the object names unchanged. For this task, we employ an open-source large language model, Llama-3.1-70B~\cite{llama3.1}, to generate 10 negative samples for each positive sample. To ensure clarity, we remove special symbols \bw{such as semicolons, commas, and line breaks from the generated descriptions.} A quality check of 3,000 negative samples reveals that 98.9\% are qualified, with only 1.1\% considered noise—a level within the expected tolerance for unsupervised methods. This process generates subtle variations that better reflect real-world scenarios where objects may appear similar but differ in specific details. We illustrate examples of the fine-grained negative samples in Appendix \ref{PNSample}.

The resulting dataset includes 12 million images with fine-grained captions, 40 million bounding boxes with detailed region descriptions, and 10 million hard negative samples. The data pipeline utilizes a cluster of 160$\times$910B NPUs and takes 7 days to complete. This comprehensive dataset enhances the model's ability to capture fine-grained details and provides a robust foundation for training FG-CLIP to distinguish subtle differences in visual and textual features.



\section{Experiments}
\label{experiments}

\subsection{Implementation Details}
In the first stage, we train on a dataset of 1.6 billion \xie{images, each paired with short and long texts}. The model is initialized with weights from the original CLIP \cite{openaiclip}. For both ViT-B and ViT-L \xie{\cite{vit} configurations, the batch size per NPU is set to 384. The learnable temperature parameter $\tau$ is initialized to 0.07. We utilize} the AdamW optimizer with a learning rate of 1e-4, weight decay of 0.05, $\beta_{1}$ of 0.9, $\beta_{2}$ of 0.98, and warmup steps \xie{for} the first 200 iterations. The entire training process employs DeepSpeed's Zero-2 optimization technique and Bfloat16 precision to accelerate training, and the model is trained for one epoch.

\begin{table}[!tbp]
\caption{R\TODO{esults on FG-OVD benchmark. Accuracy is reported}.}
\label{fgovd}
\vskip 0.1in
\begin{center}
\begin{small}
\scalebox{0.99}{
\begin{tabular}{lccccc}
\toprule
\multirow{2}{*}{Method} & \multirow{2}{*}{Backbone} & \multicolumn{4}{c}{Fine-Grained Understanding} \\
&&hard&medium&easy&trivial \\
\midrule
CLIP&ViT-B/16&12.0&23.1&22.2&58.5\\
EVA-CLIP&ViT-B/16&14.0&30.1&29.4&58.3\\
Long-CLIP&ViT-B/16&9.2&18.4&16.2&51.8\\
FineCLIP&ViT-B/16&26.8&49.8&50.4&71.9\\
FG-CLIP&ViT-B/16&\textbf{46.1}&\textbf{66.6}&\textbf{68.7}&\textbf{83.4}
\\
\midrule
CLIP&ViT-L/14&15.4&25.3&25.7&38.8\\
EVA-CLIP&ViT-L/14&18.3&38.4&35.2&62.7\\
Long-CLIP&ViT-L/14&9.6&19.7&16.0&39.8\\
FineCLIP&ViT-L/14&22.8&46.0&46.0&73.6\\
FG-CLIP&ViT-L/14&\textbf{48.4}&\textbf{69.5}&\textbf{71.2}&\textbf{89.7}\\
\bottomrule
\end{tabular}
}
\end{small}
\end{center}
\vskip -0.2in
\end{table}

\begin{table}[t]
\caption{\TODO{Bounding box classification results.}}
\label{bbc}
\begin{center}
\begin{small}
\scalebox{0.99}{
\begin{tabular}{lcccc}
\toprule
\multirow{2}{*}{Method} & \multirow{2}{*}{Backbone} & \multicolumn{3}{c}{BBox Classification} \\
&&COCO&LVIS&Open Images \\
\midrule
CLIP&ViT-B/16&44.2&20.9&15.3\\
EVA-CLIP&ViT-B/16&30.6&14.4&8.8\\
RegionCLIP&ViT-B/16&40.0&22.2&19.1\\
CLIPSelf&ViT-B/16&43.7&7.8&11.4\\
Long-CLIP&ViT-B/16&36.7&18.2&14.9\\
FineCLIP&ViT-B/16&48.4&23.3&18.1\\
FG-CLIP&ViT-B/16&\textbf{52.3}&\textbf{28.6}&\textbf{20.6}\\
\midrule
CLIP&ViT-L/14&33.8&9.3&8.3\\
EVA-CLIP&ViT-L/14&32.1&18.3&9.3\\
Long-CLIP&ViT-L/14&35.6&10.4&8.9 \\
FineCLIP&ViT-L/14&54.5&22.5&19.1\\
FG-CLIP&ViT-L/14&\textbf{63.2}&\textbf{38.3}&\textbf{23.8}\\
\bottomrule
\end{tabular}
}
\end{small}
\end{center}
\vskip -0.2in
\end{table}

In the second stage, we train on a dataset of \xie{12 million images. Apart from long and short captions, this dataset includes high-quality visual grounding annotations and hard fine-grained negative samples. The model is initialized with weights obtained from the first stage. }The batch size per GPU is set to 512. We employ the AdamW optimizer with a learning rate of 1e-6, weight decay of 0.001, $\beta_{1}$ of 0.9, $\beta_{2}$ of 0.98, and warmup steps \xie{for} the first 50 iterations. \xie{Training acceleration techniques include DeepSpeed's Zero-2 optimization, CUDA's TF32 technology, and Bfloat16 precision, and the model is trained for one epoch.}

\subsection{Comparisons on Fine-grained Region-level Task}
\label{exp_region}
\bw{
In this section, the primary methods included for comparison are CLIP \cite{openaiclip}, EVA-CLIP \cite{sun2023eva}, Long-CLIP \cite{zhang2025long}, and FineCLIP \cite{jingfineclip}. Additional methods involved in open-vocabulary detection include OV-RCNN \cite{zareian2021open}, RegionCLIP \cite{zhong2022regionclip}, Detic \cite{zhou2022detecting}, VLDet \cite{lin2022learning}, RO-ViT \cite{kim2023region}, CFM-ViT \cite{kim2023contrastive}, F-ViT\cite{wu2023clipself}, and CLIPSelf \cite{wu2023clipself}.
}


         

  

\paragraph{Fine-Grained Understanding.}

\begin{table}[!tbp]
\caption{Performance on open-vocabulary object detection task.}
\label{ovd}
\begin{center}
\vskip -0.1in
\begin{small}
\scalebox{0.95}{
\begin{tabular}{lcc>{\color{gray!70}}c>{\color{gray!70}}c}
\toprule
\multirow{2}{*}{Method} & \multirow{2}{*}{Backbone} & \multicolumn{3}{c}{OV-COCO} \\
&&$AP_{50}^{novel}$&$AP_{50}^{base}$&$AP_{50}^{all}$ \\
\midrule
OV-RCNN&RN50&17.5&41.0&34.9\\
RegionCLIP&RN50&26.8&54.8&47.5\\
Detic&RN50&27.8&51.1&45.0\\
VLDet&RN50&32.0&50.6&45.8\\
RO-ViT&ViT-B/16&30.2&-&41.5\\
RO-ViT&ViT-L/16&33.0&-&47.7\\
CFM-ViT&ViT-L/16&34.1&-&46.0\\
\midrule
F-ViT&ViT-B/16&17.5&41.0&34.9\\
F-ViT+CLIPSelf&ViT-B/16&33.6&54.2&48.8\\
F-ViT+FineCLIP&ViT-B/16&29.8&45.9&41.7\\
F-ViT+FG-CLIP&ViT-B/16&\textbf{35.1}&51.7&47.4\\
\midrule
F-ViT&ViT-L/14&24.7&53.6&46.0\\
F-ViT+CLIPSelf&ViT-L/14&38.4&60.6&54.8\\
F-ViT+FineCLIP&ViT-L/14&40.0&57.2&52.7\\
F-ViT+FG-CLIP&ViT-L/14&\textbf{41.2}&58.0&53.6\\

\bottomrule
\end{tabular}
}
\end{small}
\end{center}
\vskip -0.2in
\end{table}

\begin{table*}[!tbp]
\caption{Comparisons on image-level tasks, including long/short caption image-text retrieval, and zero-shot image classification. }
\label{sl_it}
\vskip 0.1in
\begin{center}
\begin{small}
\scalebox{1.0}{
\begin{tabular}{lccccccccccc}
\toprule
\multirow{2}{*}{Method} & \multirow{2}{*}{Backbone} 
& \multicolumn{2}{c}{ShareGPT4V} & \multicolumn{2}{c}{DCI} 
& \multicolumn{2}{c}{MSCOCO}& \multicolumn{2}{c}{Flickr30k}
&ImageNet-1K&ImageNet-v2\\
&&I2T&T2I&I2T&T2I&I2T&T2I&I2T&T2I&Top-1&Top-1 \\
\midrule
CLIP&ViT-B/16&78.2&79.6&45.5&43.0&51.8&32.7&82.2& 62.1&68.4&61.9\\
EVA-CLIP&ViT-B/16&90.5&85.5&41.9&41.2&58.7&41.6&85.7&71.2&\textbf{74.7}&\textbf{67.0}\\
Long-CLIP&ViT-B/16&94.7&93.4&51.7&57.3&57.6&40.4&85.9&70.7 &66.8&61.2\\
FineCLIP&ViT-B/16&70.6&73.3&35.5&34.4&54.5&40.2&82.5&67.9&55.7&48.8\\
FG-CLIP&ViT-B/16&\textbf{96.7}&\textbf{94.9}&\textbf{61.8}&\textbf{60.6}&\textbf{64.1}&\textbf{45.4}&\textbf{90.7}&\textbf{76.4}&69.0&61.8\\
\midrule
CLIP&ViT-L/14&86.5&83.6&37.2&36.4&58.0&37.1&87.4&67.3&76.6&70.9\\
EVA-CLIP&ViT-L/14&91.5&89.4&47.2&47.8&64.2&47.9&89.2&77.9&\textbf{80.4}&\textbf{73.8}\\
Long-CLIP&ViT-L/14&95.8&95.6&44.2&52.5&62.8&46.3&90.0&76.2&73.5&67.9\\
FineCLIP&ViT-L/14&73.4&82.7&40.1&46.2&-&-&-&-&60.8&53.4\\
FG-CLIP&ViT-L/14&\textbf{97.4}&\textbf{96.8}&\textbf{66.7}&\textbf{66.1}&\textbf{68.9}&\textbf{50.9}&\textbf{93.7}&\textbf{81.5}&76.1&69.0\\
\bottomrule
\end{tabular}

}
\end{small}
\end{center}
\vskip -0.2in
\end{table*}

Based on the fine-grained benchmark FG-OVD constructed by \cite{bianchi2024devil}, we evaluate open-source image-text alignment models. Unlike previous benchmarks such as MSCOCO \cite{lin2014microsoft} and Flickr \cite{young2014image}, which rely on global information for matching, this benchmark focuses on identifying specific local regions within images. Each region has one corresponding positive description and ten negative descriptions, with the negative samples derived from the positive text. This benchmark primarily comprises four subsets of varying difficulty levels: hard, medium, easy, and trivial. The increasing difficulty across these subsets is reflected in the degree of distinction between the texts to be matched. In the hard, medium, and easy subsets, one, two, and three attribute words are replaced, respectively. In the trivial subset, the texts are entirely unrelated. For the source collection of specific attribute words, please refer to \cite{bianchi2024devil}.


During testing, following FineCLIP, we first extract dense features from the model by removing the last self-attention layer as suggested by \cite{zhou2022extract}. 
Subsequently, we combine the bounding box information provided by the benchmark with ROIAlign to obtain representative features. These features are used to calculate similarity scores with both positive and negative sample descriptions. Top-1 accuracy is adopted as the evaluation metric.

\xie{As shown in Table \ref{fgovd}, FG-CLIP achieves significant improvements over existing models, particularly on the challenging hard and medium subsets, thanks to its hard fine-grained negative samples learning strategy.} \TODO{Examples of different models' performance can be found in Appendix \ref{VC2}.}

\paragraph{Bounding Box Classification.}

\xie{To assess the model's local information recognition capabilities, \TODO{we conduct zero-shot testing on COCO-val2017~\cite{lin2014microsoft}, LVIS~\cite{gupta2019lvis}, and Open Images~\cite{openimage}, following the protocol of~\cite{jingfineclip}}. This evaluation focuses on how well the model can classify objects within bounding boxes using only textual descriptions. Similar to the fine-grained understanding, we integrate known bounding box information from the benchmark with ROIAlign to obtain localized dense representations. Using all categories as textual inputs, we perform matching and recognition for each bounding box, evaluating \TODO{Top-1 accuracy.}}


As shown in Table \ref{bbc}, FG-CLIP achieves leading performance in bounding box classification with the help of the regional contrastive learning strategy. Notably, Long-CLIP~\cite{zhang2025long}, fine-tuned from CLIP using long texts, shows a significant decline in performance, indicating that long texts affect regional information granularity.
Furthermore, FineCLIP uses region alignment data and incorporates a real-time self-distillation scheme, leading to meaningful improvements. While FineCLIP makes significant progress, FG-CLIP excels it by integrating regional and global information. This approach enhances FG-CLIP's ability to accurately recognize and classify regions within images, highlighting the effectiveness of FG-CLIP's training strategy.


\vskip -0.2in
\paragraph{Open-Vocabulary Object Detection.} To further evaluate the fine-grained localization capability of our method, we employ FG-CLIP as the backbone for downstream open-vocabulary detection tasks. Following prior work \cite{wu2023clipself}, we employ a two-stage detection architecture, F-VIT, with a frozen visual encoder. The comparative results are summarized in Table \ref{ovd}. Consistent with previous studies, we report the box AP at IoU 0.5 for base, novel, and all categories ($AP_{50}^{novel}$, $AP_{50}^{base}$, and $AP_{50}^{all}$) on OV-COCO. Notably, $AP_{50}^{novel}$ is the primary focus of interest, as it measures the model's ability to recognize novel objects. Our findings indicate that FG-CLIP achieves leading performance in open-vocabulary detection tasks, highlighting its effectiveness in recognizing and localizing novel objects.

\subsection{Comparisons on Image-level Task}
\paragraph{Long/short Caption Image-Text Retrieval.}
\bw{
\xie{To evaluate retrieval performance comprehensively, we conduct experiments on both long caption and short caption image-text retrieval tasks. For long-text retrieval, we follow the protocol of Long-CLIP and use the 1K subset of ShareGPT4V \cite{chen2025sharegpt4v} provided by it as the testset.} Additionally, we incorporate a more challenging long caption image-text pair dataset from DCI \cite{urbanek2024picture}, consisting of 7,805 pairs, into the evaluation. For short-text retrieval, we employ the classic \xie{MSCOCO 5K~\cite{lin2014microsoft} and Flickr 1K~\cite{young2014image} evaluation sets, which are widely used benchmarks for assessing image-text alignment models.} As shown in Table \ref{sl_it}, FG-CLIP achieves significant performance improvements in both long/short caption image-text retrieval tasks. \xie{The model's ability to handle diverse caption lengths highlights its versatility and robustness in multimodal alignment.}}

\begin{table}[!tbp]
\caption{Comparisons on General Multimodal
Benchmarks.}
\label{lmm}
\begin{center}
\vskip -0.1in
\begin{small}
\scalebox{0.92}{
\begin{tabular}{lcccccc}
\toprule
\multirow{2}{*}{Method} & \multirow{2}{*}{GQA} & \multirow{2}{*}{POPE} &\multicolumn{3}{c}{RefCOCO} \\
&&&val&testA&testB \\
\toprule  
LLaVA-v1.5+CLIP&61.9&85.9&76.2&83.4&67.9\\
&{\textit{+1.0}}&{\textit{+0.9}}&{\textit{+5.2}}&{\textit{+3.1}}&{\textit{+7.0}}\\
LLaVA-v1.5+FG-CLIP&62.9&86.8&81.4&86.5&74.9\\

\bottomrule
\end{tabular}
}
\end{small}
\end{center}
\vskip -0.2in
\end{table}

\begin{table*}[t]
\vskip -0.1in
  \centering
  \small              
  \caption{\TODO{Ablation study results for FG-CLIP. This table compares the performance of different configurations of our FG-CLIP model across multiple evaluation metrics, including long caption image-text retrieval (DCI), short caption image-text retrieval (MSCOCO), bounding box classification (COCO-val2017), and fine-grained understanding (FG-OVD). The results highlight the incremental improvements achieved by incorporating global contrastive learning ($L_{global}$), regional contrastive learning ($L_{regional}$), and hard fine-grained negative samples learning ($L_{hard}$).}}
  \vskip 0.1in
  \begin{tabular}{lcccccccccc}
  \toprule           
         \multirow{2}{*}{Method} & 
         \multicolumn{2}{c}{Long Retrieval}&
         \multicolumn{2}{c}{Short Retrieval} & 
         \multicolumn{2}{c}{BBox Classification} & 
         \multicolumn{3}{c}{Fine-Grained Understanding} & \\
&I2T&T2I&I2T&T2I&\multicolumn{1}{c}{Top-1}&Top-5&hard&medium&easy&\\
\midrule
    CLIP   &
    
    45.5&43.0&51.8&32.7&44.2&72.3&12.0&23.1&22.2\\
    \midrule
    FG-CLIP Stage1   & 
    58.3&57.5&64.6&44.9&47.2&74.2&21.8&41.6&36.2\\
     ~+Stage2 ($L_{global}$)   & 
    62.7&61.2&64.4&46.4&46.8&73.6&25.4&46.8&42.9\\
     ~+Stage2 ($L_{global}$,$L_{regional}$)   &  
    62.4&61.1&64.7&45.7&53.7&81.2&24.5&47.1&49.5\\
    ~+Stage2 ($L_{global}$,$L_{regional}$,$L_{hard}$)   & 
     61.8&60.6&64.1&45.4&52.3&79.7&46.1&66.6&68.7\\
     \bottomrule
  \end{tabular}
  
   \label{tab:aba_strc}
   \vskip -0.1in
\end{table*}

\paragraph{Zero-shot Image Classification.}

\bw{
We evaluate the zero-shot classification performance of our model on ImageNet-1K \cite{deng2009imagenet} and ImageNet-v2 \cite{recht2019imagenet}. As illustrated in Table \ref{sl_it}, \xie{despite being marginally behind EVA-CLIP, which is trained on a larger dataset, FG-CLIP demonstrates stable classification performance with enhanced regional and textual understanding capabilities compared to the original baseline, CLIP.} Additionally, when compared to Long-CLIP and FineCLIP, both of which aim to enhance fine-grained recognition capabilities, our model exhibits a notable advantage in classification accuracy.
}

\subsection{Comparisons on General Multimodal Benchmarks} 
\bw{
We compare FG-CLIP as a visual feature extractor for multimodal large language models with our baseline, CLIP. Specifically, we conduct experiments using LLaVA-v1.5-7B \cite{llava}, which itself is trained using CLIP. To ensure a fair comparison, all parameter configurations are kept consistent with those in the original LLaVA, and the model is trained using the data provided by LLaVA. \xie{Our evaluation focuses on benchmark sets} related to attribute analysis, object localization, and output hallucination, which are GQA \cite{hudson2019gqa}, RefCOCO \cite{refcoco}, and POPE \cite{pope}, respectively.
}

\bw{
As shown in Table \ref{lmm}, FG-CLIP achieves certain improvements on GQA, which \xie{involves} attribute-based question answering, and on POPE, which \xie{evaluates} output hallucination. Additionally, it demonstrates significant gains on RefCOCO, a benchmark set that involves both attribute analysis and object localization. These results indicate the effectiveness of FG-CLIP's training strategy and the data construction, which are \xie{specifically designed to enhance} fine-grained recognition and regional alignment.
} \TODO{We provide more results in Section \ref{addition:mllm}.}

\subsection{Ablation Study}
\label{Sec: ablation}
To systematically evaluate the contributions of different components in our FG-CLIP model, we conduct an ablation study with results summarized in Table \ref{tab:aba_strc}.

\paragraph{Global Contrastive Learning and Detailed Recaptioning Data.}
We start by comparing the original CLIP model with FG-CLIP Stage 1 and Stage 2 incorporating global contrastive learning $L_{global}$. The results demonstrate that generating detailed captions significantly enhances performance across various tasks. Specifically, FG-CLIP Stage 1 outperforms CLIP in all metrics, highlighting the importance of fine-grained training data. Further improvements are observed when adding $L_{global}$ in Stage 2, particularly in long caption image-text retrieval (DCI~\cite{urbanek2024picture}) and fine-grained understanding (FG-OVD~\cite{bianchi2024devil}). This underscores the effectiveness of detailed caption data combined with global contrastive learning in improving model performance.
\paragraph{Regional Contrastive Learning.}
We introduce regional contrastive learning $L_{regional}$ to evaluate its impact on capturing detailed image regions. Compared to configurations using only $L_{global}$, adding $L_{regional}$ leads to substantial improvements in \TODO{bounding box classification accuracy from 46.8\% to 53.7\%, and FG-OVD easy dataset accuracy from 42.9\% to 49.5\%}. These gains highlight the effectiveness of $L_{regional}$ in refining the model's ability to understand fine-grained details within specific image regions. Moreover, this component maintains strong performance in both retrieval and classification tasks, demonstrating its versatility.

\paragraph{Hard Fine-Grained Negative Samples Learning.}
We incorporate hard fine-grained negative samples learning $L_{hard}$ to distinguish subtle differences in semantically
similar but distinct region-text pairs. By comparing configurations with and without $L_{hard}$, we observe significant improvements in FG-OVD performance. \TODO{Accuracy on the hard dataset increases from 24.5\% to 46.1\%, while on the medium dataset it rises from 47.1\% to 66.6\% and on the easy dataset it jumps from 49.5\% to 68.7\%.} These results underscore the importance of $L_{hard}$ in distinguishing subtle semantic differences. Hard fine-grained negative samples learning effectively addresses challenge cases, thereby enhancing the model's stability and discriminative power.

\section{Conclusion}
\label{conclusion}
In this work, we introduce Fine-Grained CLIP (FG-CLIP), a novel approach that significantly advances fine-grained understanding. By integrating advanced alignment techniques with large-scale, high-quality datasets and hard negative samples, FG-CLIP captures global-level and region-level semantic details and distinguishes subtle differences more effectively. Extensive experiments across diverse downstream tasks validate the model's superior performance. In addition, we propose FineHARD as a unified dataset that combines high-quality region-specific annotations with challenging fine-grained negative samples, offering a valuable resource for advancing multimodal research. Looking ahead, exploring the integration of more sophisticated multimodal models and expanding dataset diversity will be crucial for pushing the boundaries of fine-grained understanding.

\section*{Impact Statement}

This paper aims to advance the field of Machine Learning, which has broad implications for society. There are many potential societal consequences of our work, none which we feel must be specifically highlighted here. 

{
    \small
    \bibliography{example_paper}
    \bibliographystyle{icml2025}
}

\newpage
\appendix
\onecolumn
\section{Examples of Curated Visual Grounding Data}
\label{TSample}
\begin{figure*}[!hbtp]
\setlength{\tabcolsep}{1pt}
\centering
\scriptsize
\resizebox{0.95\linewidth}{!}
{
\centering
\begin{tabular}{c}

\includegraphics[width=0.95\linewidth]{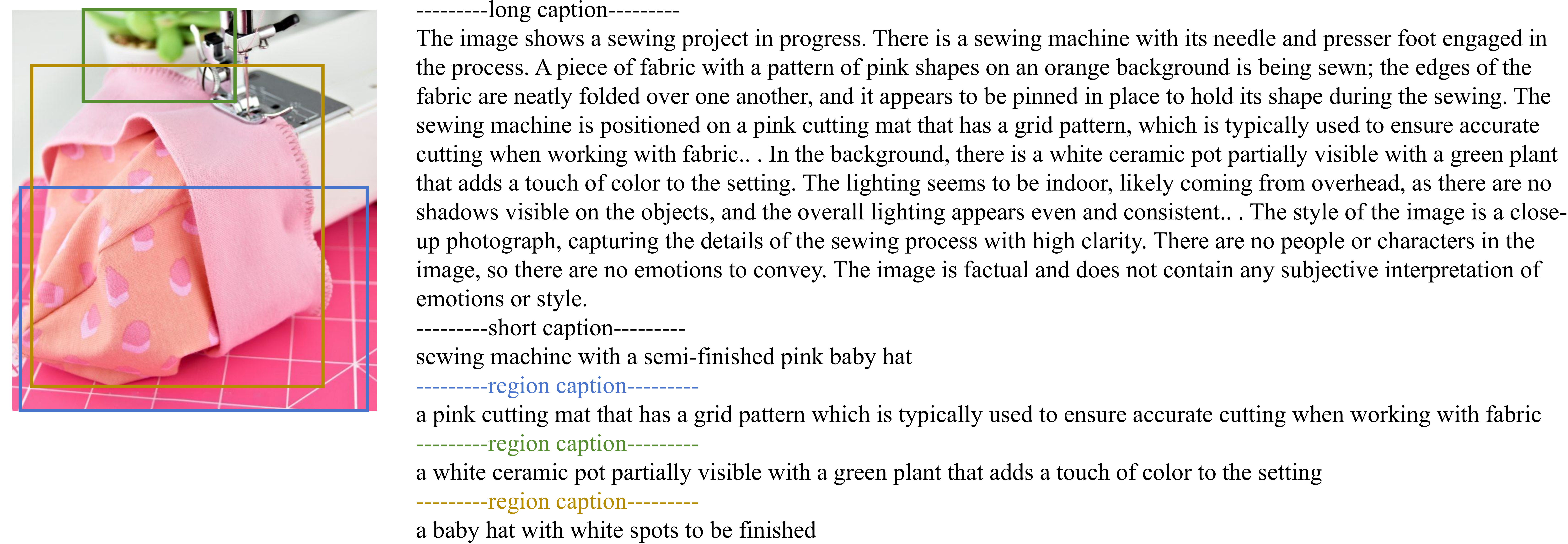}\\[10pt]
\includegraphics[width=0.95\linewidth]{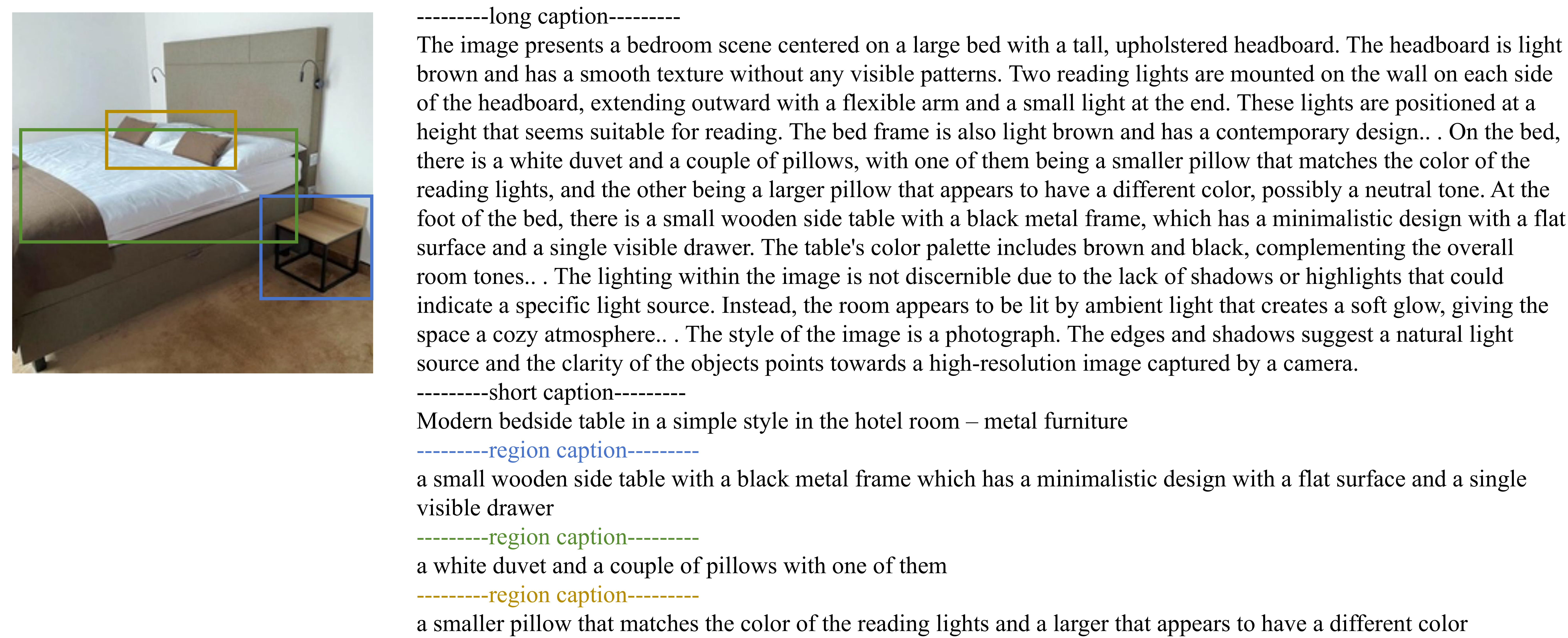}\\[10pt]
\includegraphics[width=0.95\linewidth]{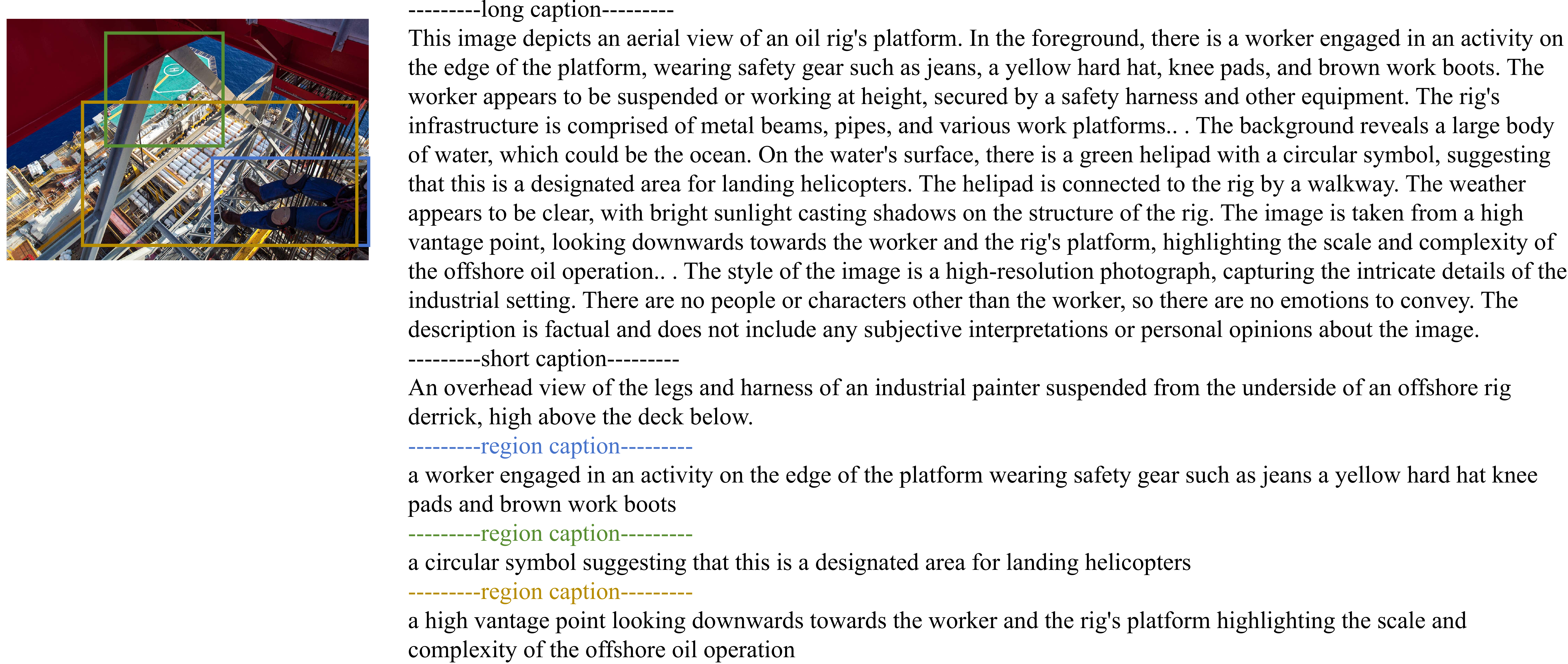}\\

\end{tabular}
}
\caption{Examples of curated visual grounding data.}
  \label{fig:traindata}
\end{figure*}

Figure \ref{fig:traindata} shows visual grounding examples utilized in our experiments. Each example comprises an image, accompanied by its corresponding long and short captions, as well as multiple region-specific annotations, each with a detailed description.

\section{Positive and Negative Descriptions Related to Image Regions}
\label{PNSample}

\begin{figure*}[!htbp]
  \centering   \includegraphics[width=0.95\linewidth]{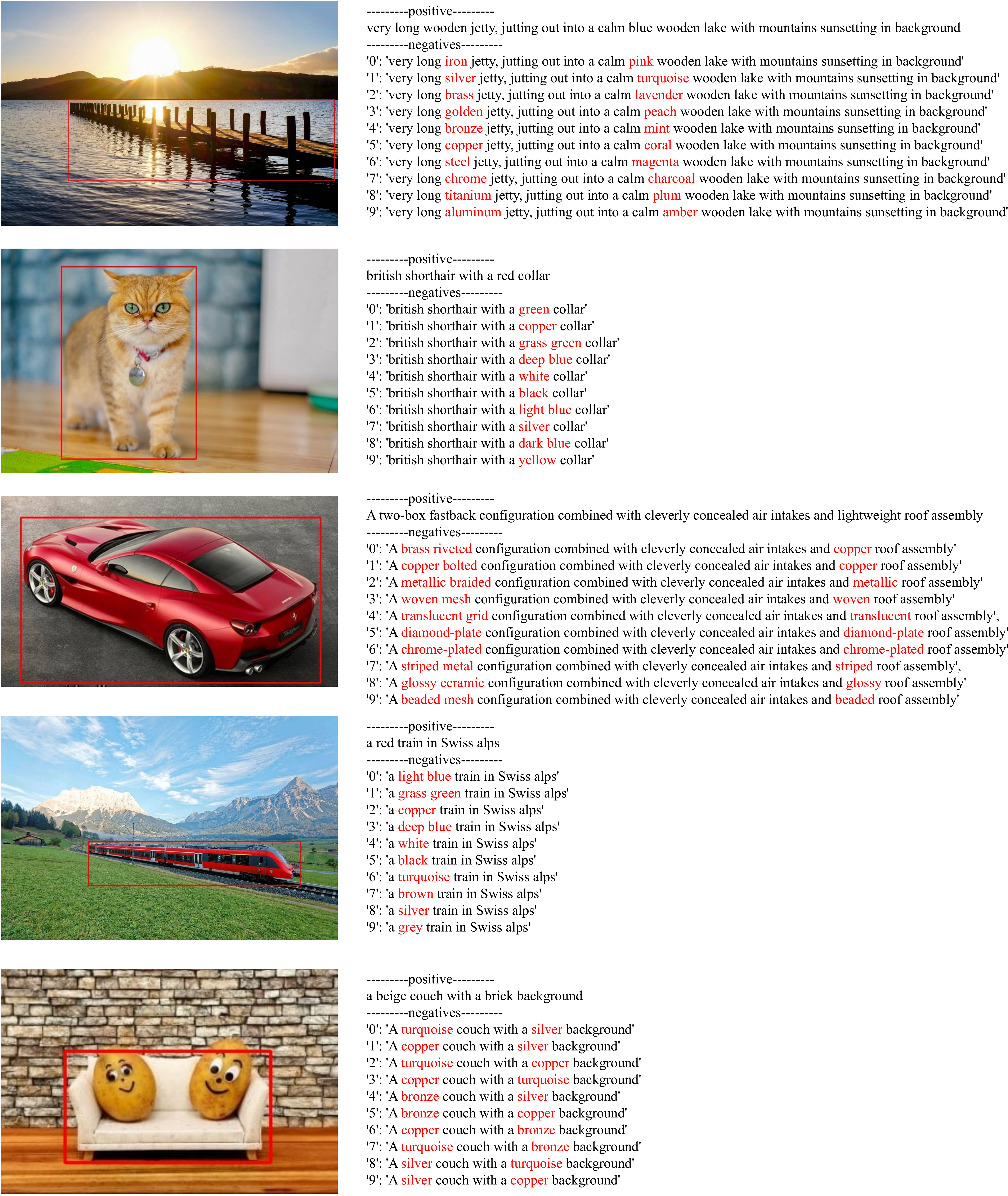}
    \caption{Examples of positive and negative descriptions related to image regions.}
    
    \label{fig:PNSample}
\end{figure*}

To generate hard fine-grained negative samples, we modify the attributes of bounding box descriptions while keeping object names unchanged. Figure \ref{fig:PNSample} illustrates examples of positive and corresponding negative descriptions for image regions.

\section{Visualization Comparison}
\label{VC}

\begin{figure*}[!hbtp]
\setlength{\tabcolsep}{1pt}
\centering
\scriptsize
\resizebox{0.92\linewidth}{!}
{
\centering
\begin{tabular}{ccccc}

\includegraphics[width=0.19\linewidth]{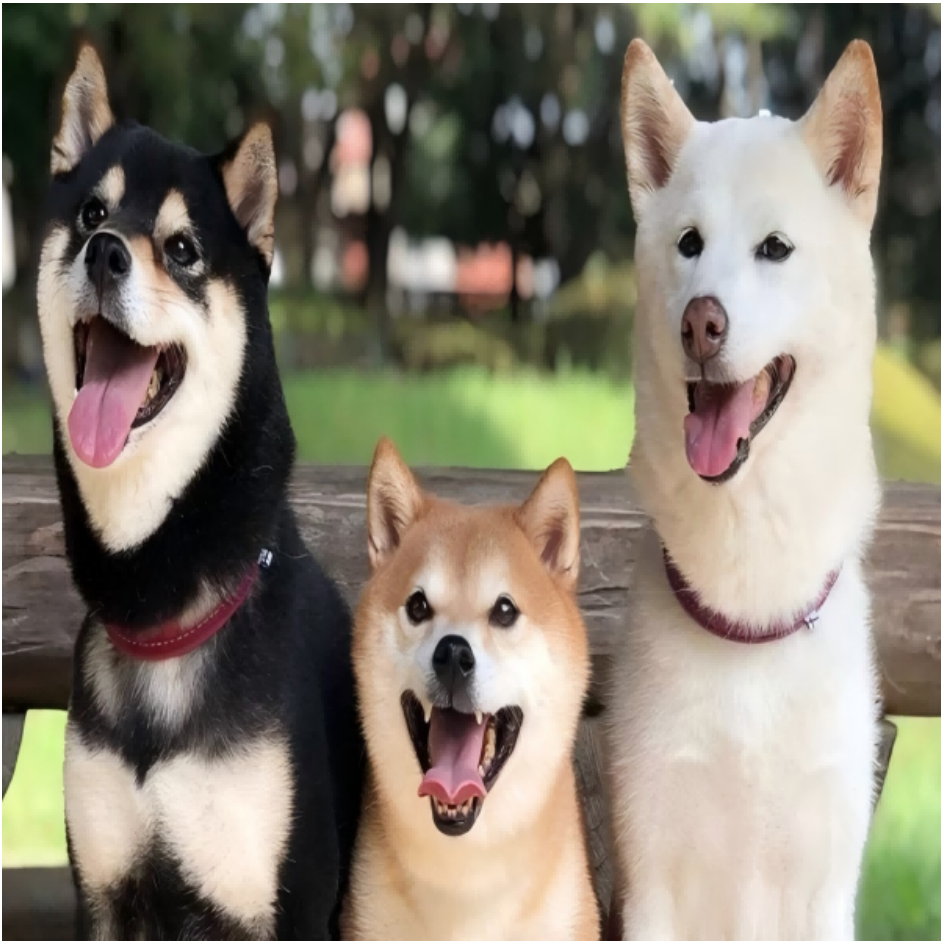}&
\includegraphics[width=0.19\linewidth]{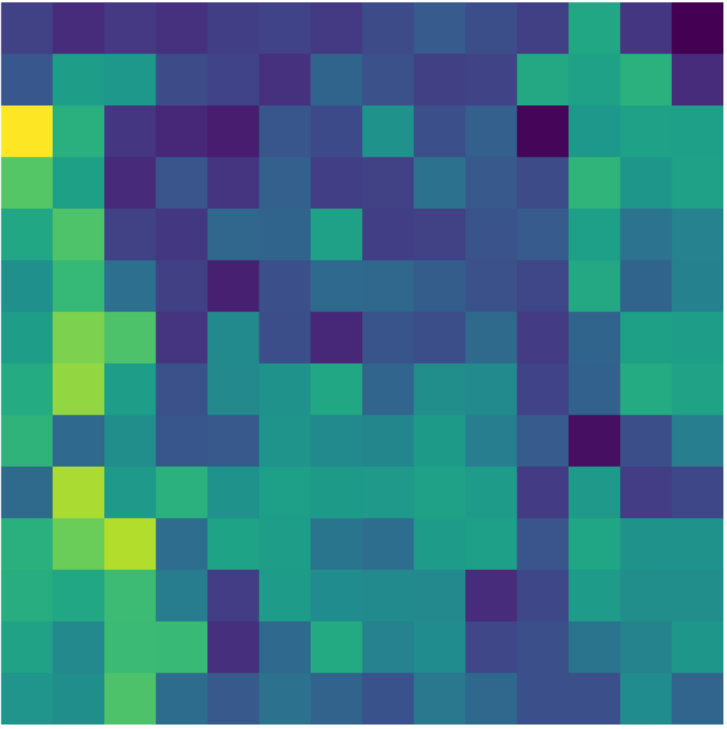}&
\includegraphics[width=0.19\linewidth]{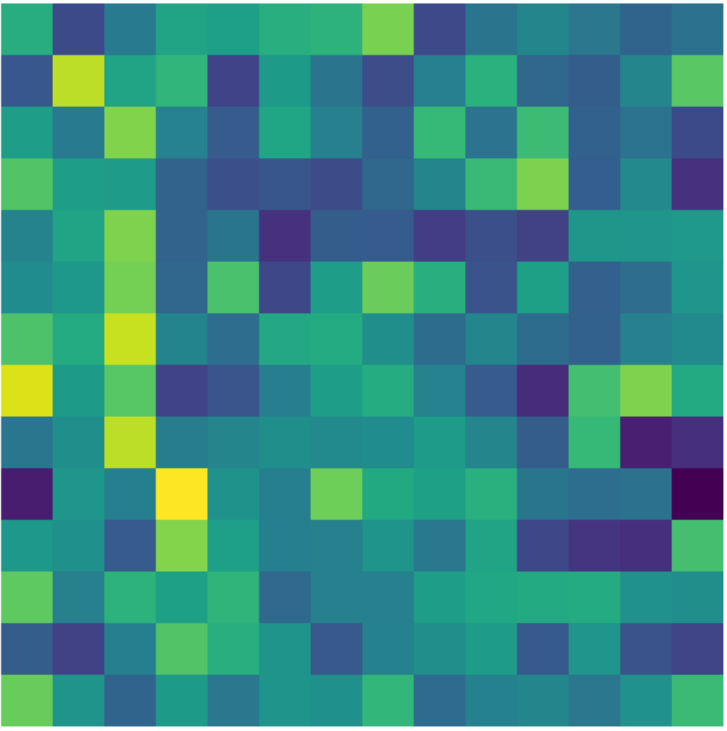}&
\includegraphics[width=0.19\linewidth]{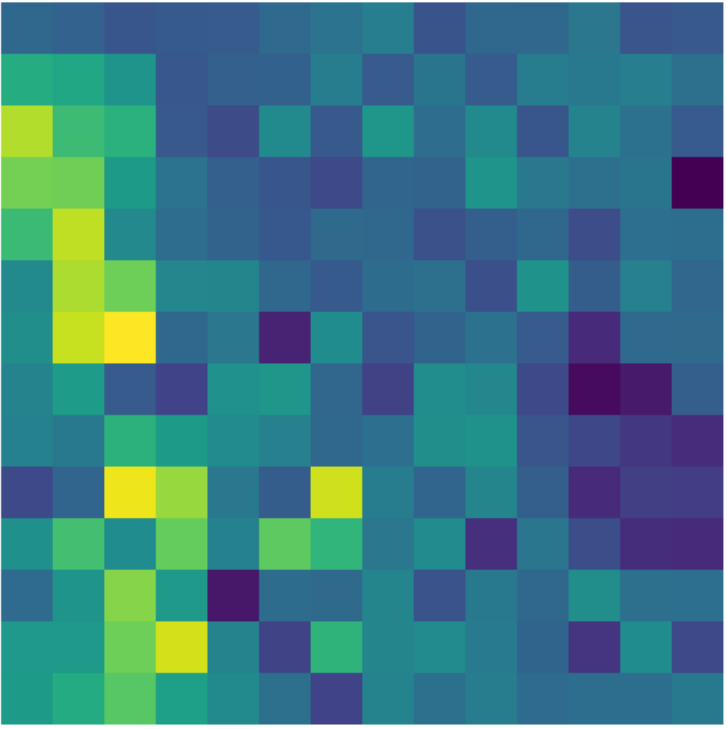}&
\includegraphics[width=0.19\linewidth]{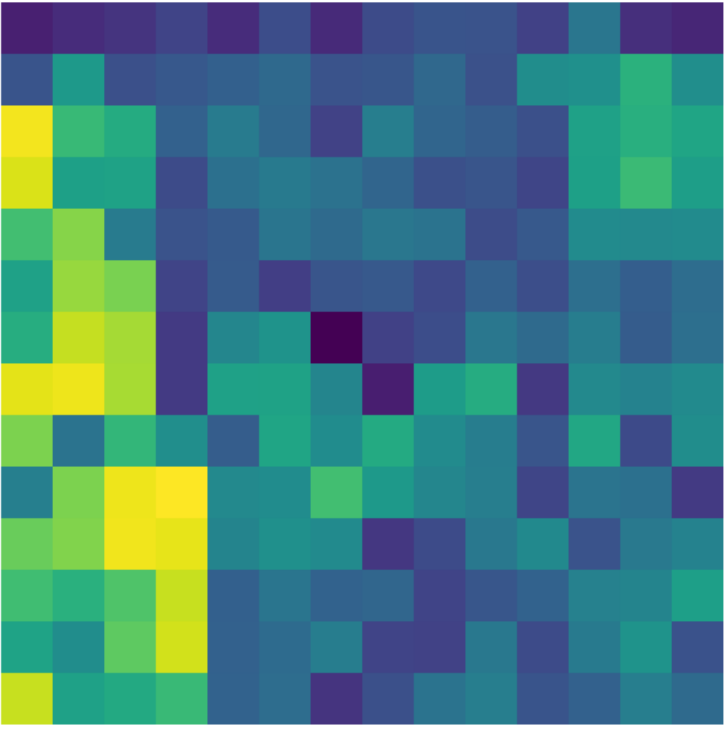}\\

Black dog & CLIP& EVA-CLIP&FineCLIP&FG-CLIP\\

\includegraphics[width=0.19\linewidth]{figures/blacknose/blacknose.pdf}&
\includegraphics[width=0.19\linewidth]{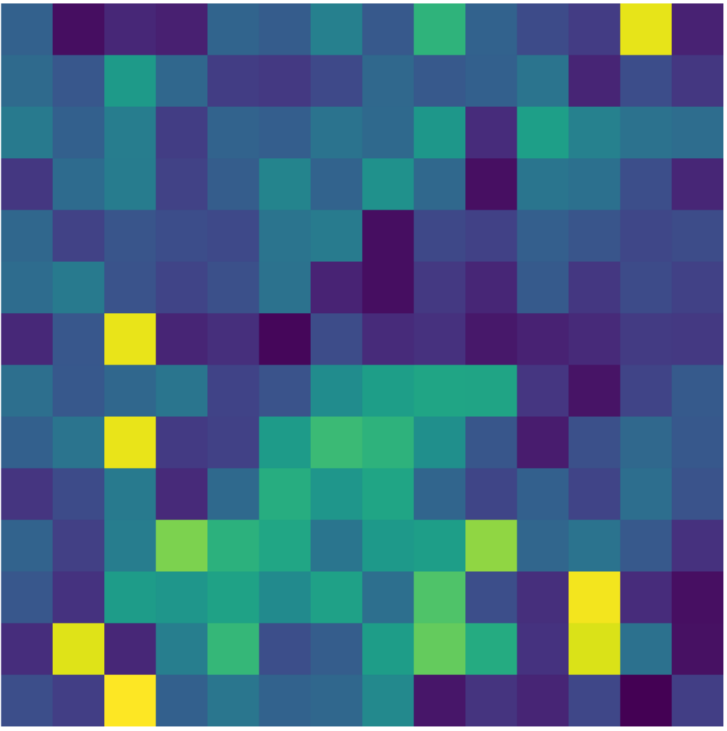}&
\includegraphics[width=0.19\linewidth]{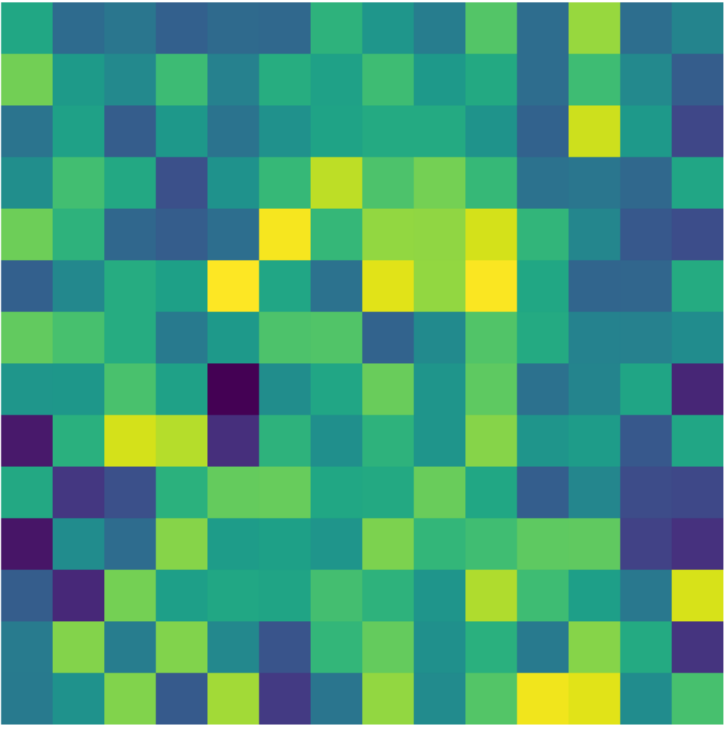}&
\includegraphics[width=0.19\linewidth]{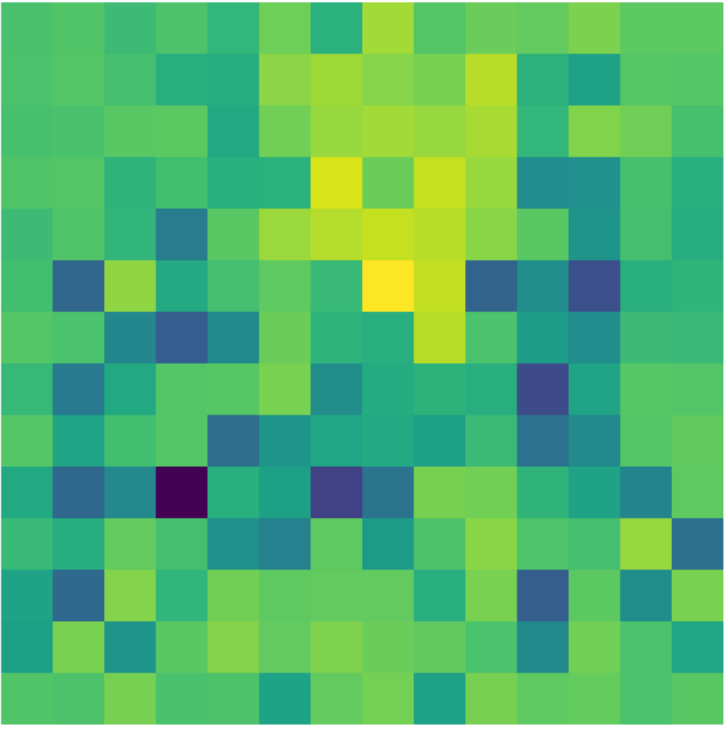}&
\includegraphics[width=0.19\linewidth]{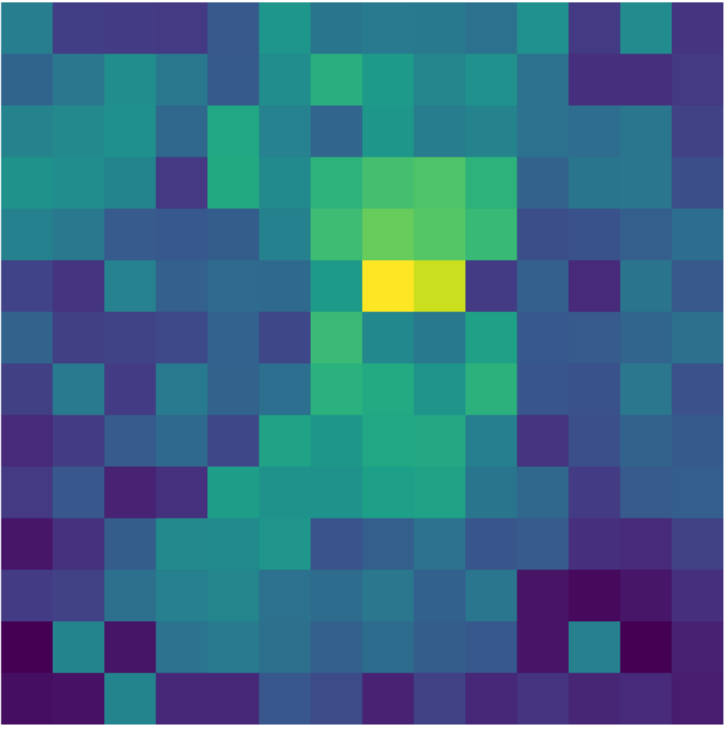}\\

Black nose & CLIP& EVA-CLIP&FineCLIP&FG-CLIP\\

\includegraphics[width=0.19\linewidth]{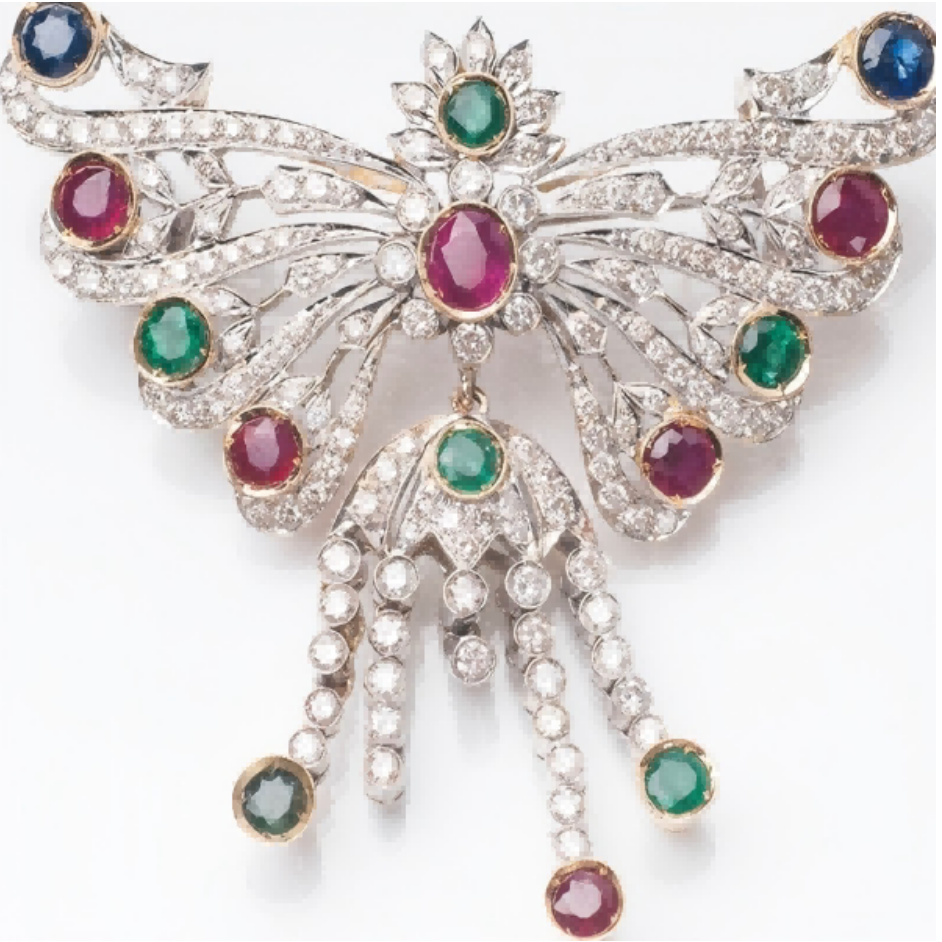}&
\includegraphics[width=0.19\linewidth]{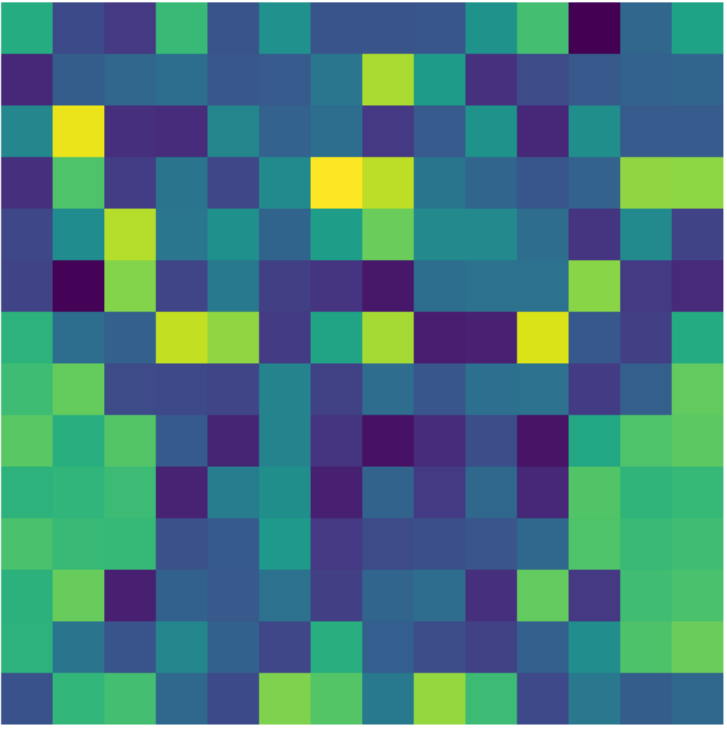}&
\includegraphics[width=0.19\linewidth]{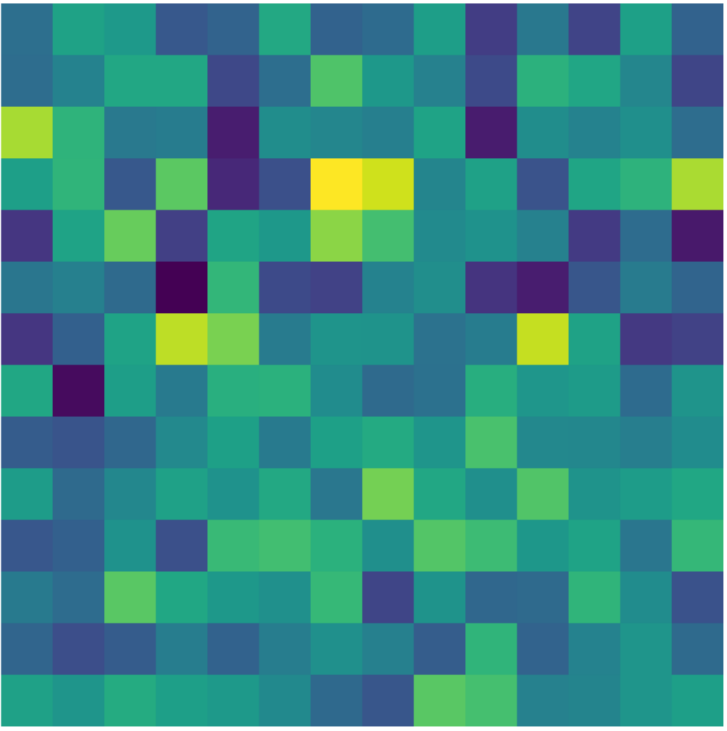}&
\includegraphics[width=0.19\linewidth]{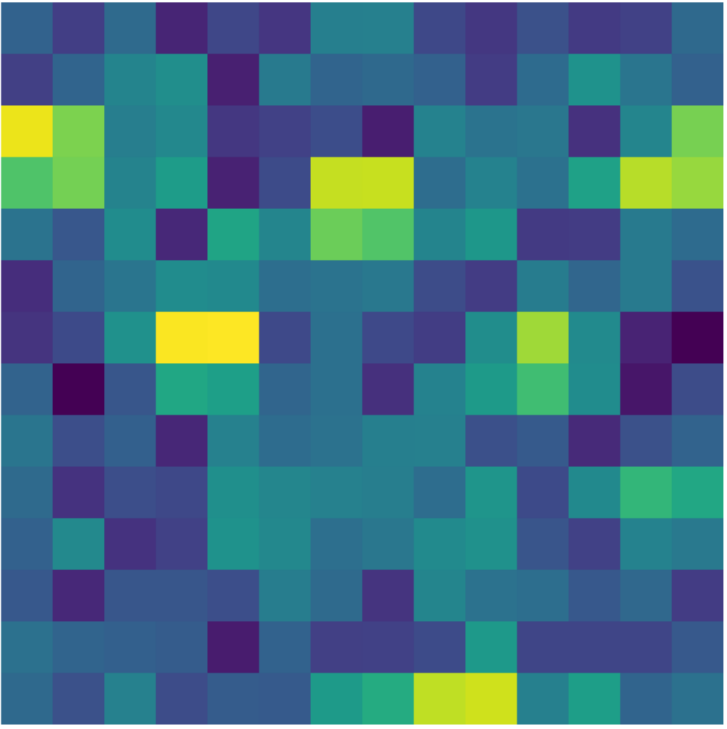}&
\includegraphics[width=0.19\linewidth]{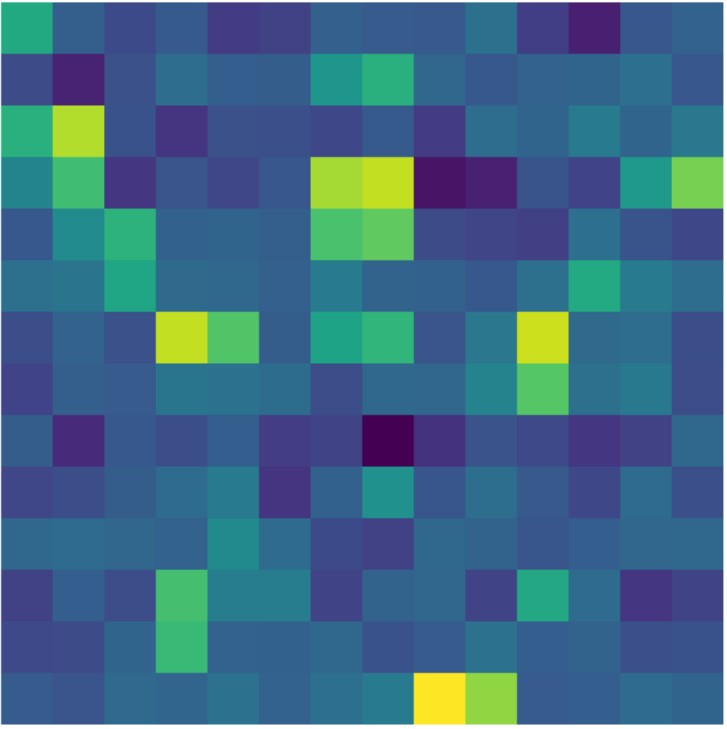}\\

Red gemstone & CLIP& EVA-CLIP&FineCLIP&FG-CLIP\\
\end{tabular}
}
\caption{Feature visual comparisons of different methods.}
  \label{fig:fvc}
\end{figure*}

As illustrated in Figure \ref{fig:fvc}, we present a comparison of similarity matrix visualizations for different methods using challenging sample images. We utilize the dense image feature extraction strategy introduced by \cite{zhou2022extract}. In the figure, warmer colors (e.g., yellow) denote higher similarity, whereas cooler colors (e.g., blue) indicate lower relevance. Our goal is for the model to precisely comprehend and interpret the fine-grained details within the images.

In the first scenario, the image contains three dogs of different colors, and we compute the similarity matrix using only the phrase "Black dog" with each image token. It can be observed that CLIP and EVA-CLIP fail to accurately identify the target dog, FineCLIP captures some relevant tokens but not all, whereas FG-CLIP identifies a larger number of relevant tokens, demonstrating superior performance.

In the second scenario, the image contains multiple black entities, and we compute the similarity matrix using only the phrase "Black nose", which occupies a very small area within the image. CLIP fails to identify the target, while EVA-CLIP and FineCLIP locate the target but also respond to many other black regions. In contrast, FG-CLIP accurately identifies the target, showcasing its precision in fine-grained localization.

In the third scenario, the image features gemstones of three different colors, and we compute the similarity matrix using only the phrase "Red gemstone". Both CLIP and EVA-CLIP fail to locate the target at the bottom and exhibit high responses to gemstones of other colors. FineCLIP shows slightly lower localization accuracy compared to FG-CLIP, which precisely distinguishes between the colors of different gemstones and achieves more accurate localization.




\begin{figure*}[!hbtp]
\setlength{\tabcolsep}{1pt}
\centering
\scriptsize
\resizebox{0.96\linewidth}{!}
{
\centering
\begin{tabular}{c}

\includegraphics[width=0.19\linewidth]{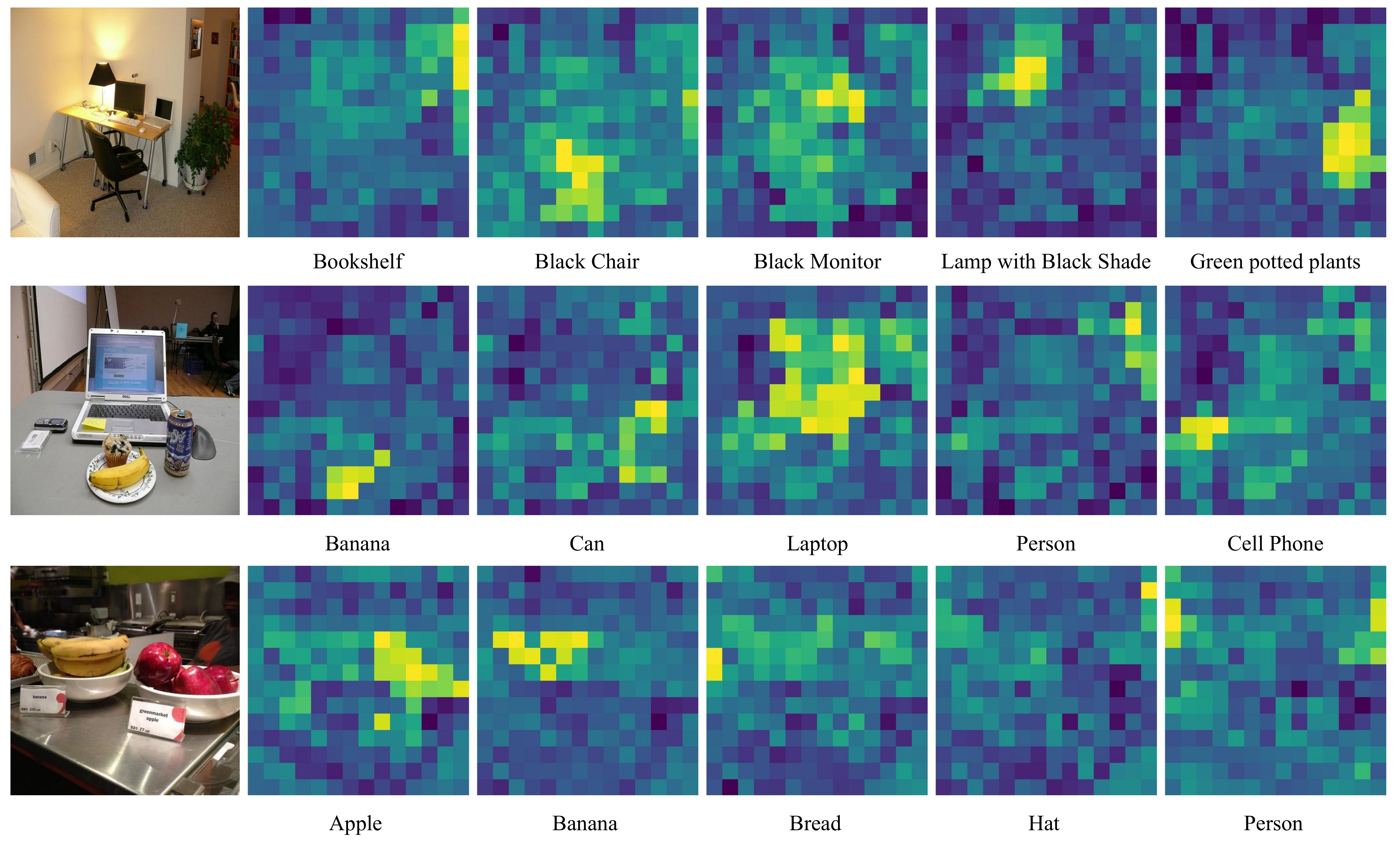}

\end{tabular}
}
\caption{Feature visual comparisons of different input texts.}
  \label{fig:part}
\end{figure*}

\TODO{Additionally, we utilize FG-CLIP to conduct a correlation analysis between different input texts and the same image. The results in Figure~\ref{fig:part} indicate that FG-CLIP provides precise positional understanding of different targets within the image. This demonstrates the model's stable visual localization capabilities and its fine-grained understanding of image content.}

\begin{figure*}[!hbtp]
\setlength{\tabcolsep}{1pt}
\centering
\scriptsize
\resizebox{0.96\linewidth}{!}
{
\centering
\begin{tabular}{ccccc}

\includegraphics[width=0.19\linewidth]{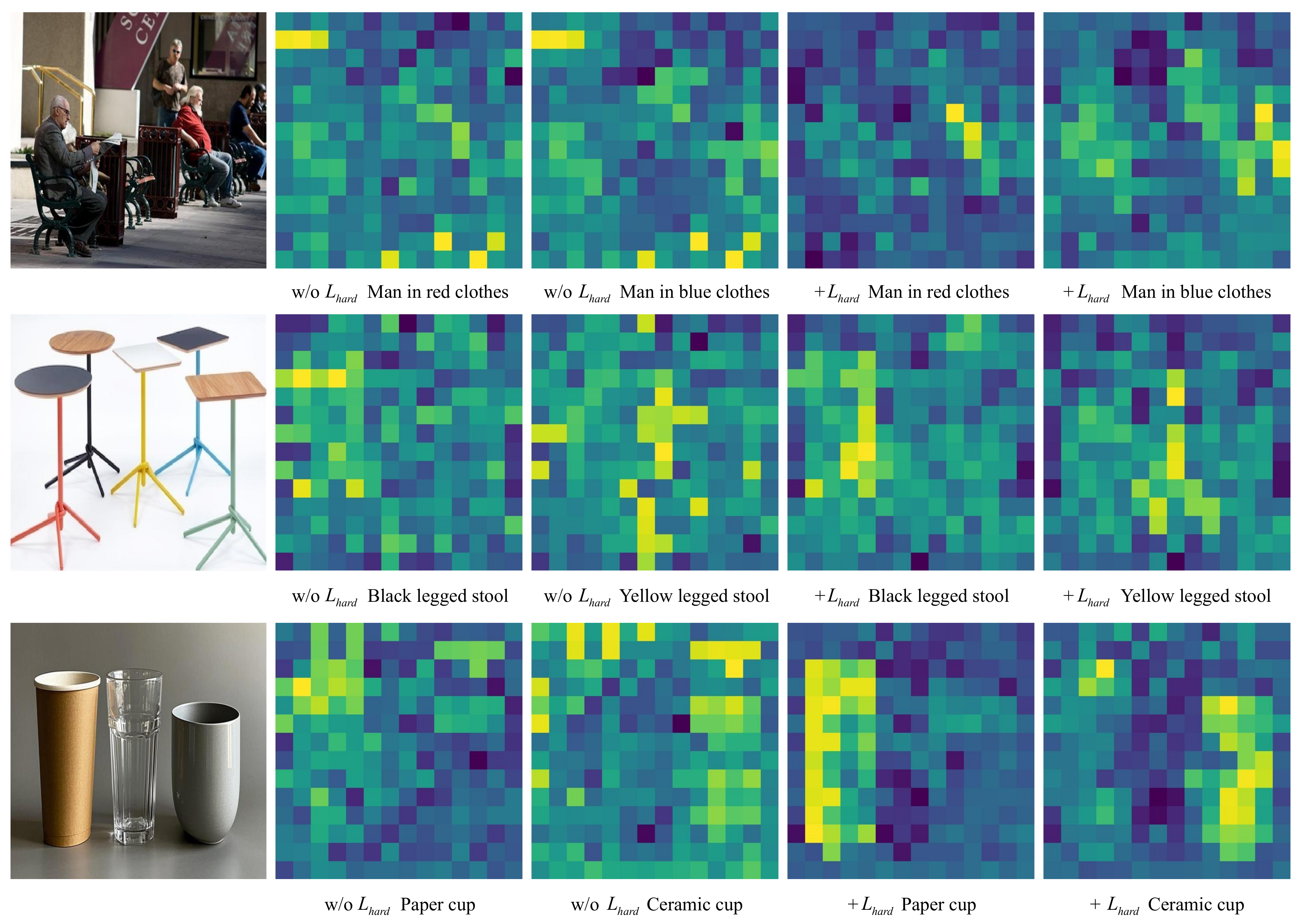}
\end{tabular}
}
\caption{Performance of hard fine-grained negative samples learning.}
  \label{fig:fg-learning}
\end{figure*}

\TODO{To evaluate the impact of hard fine-grained negative samples learning, we further provide the qualitative results in Figure \ref{fig:fg-learning}. After performing hard negative sampling, our FG-CLIP can capture the regions more accurately. For example, the highlighted region of "Man in red clothes" with hard negative loss in 1st row shows significantly better than that without hard negative loss.}


\section{Further Experiments}
\label{VD}

\begin{table*}[!htbp]
\caption{Comparisons of different methods on fine-grained benchmark.}
\label{fgvc}
\vskip 0.1in
\begin{center}
\begin{small}
\scalebox{0.88}{
\begin{tabular}{clcccc}
\toprule
Image with region&Positive and \red{Negative} Region Descriptions&CLIP&EVA-CLIP&FineCLIP&FG-CLIP\\
\toprule  
\multirow{11}{*}{\includegraphics[width=0.25\linewidth]{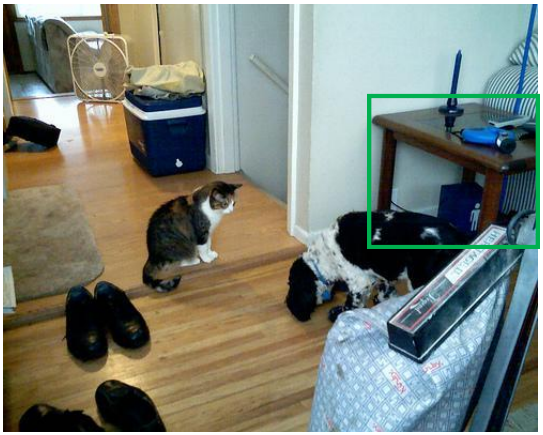}}&Origin: A table made of dark brown wood.&0.73&0.79&0.62&\green{1.0}\\
&1: A table made of \red{pink} wood.&0.0&0.59&0.58&0.0\\
&2: A table made of dark brown \red{paper}.&0.48&0.34&0.48&0.48\\
&3: A table made of \red{light grey} wood.&0.27&0.0&0.39&0.55\\
&4: A table made of dark brown \red{wool}.&0.27&0.11&0.03&0.21\\
&5: A table made of \red{yellow} wood.&0.38&0.44&0.38&0.14\\
&6: A table made of dark brown \red{velvet}.&\green{1.0}&0.63&0.0&0.01\\
&7: A table made of dark brown \red{text}.&0.94&0.72&0.38&0.55\\
&8: A table made of \red{grey} wood.&0.53&0.47&\green{1.0}&0.61\\
&9: A table made of dark brown \red{plastic}.&0.45&\green{1.0}&0.86&0.26\\
&10: A table made of \red{green} wood.&0.47&0.16&0.64&0.46\\
\midrule
\multirow{11}{*}{\includegraphics[width=0.25\linewidth]{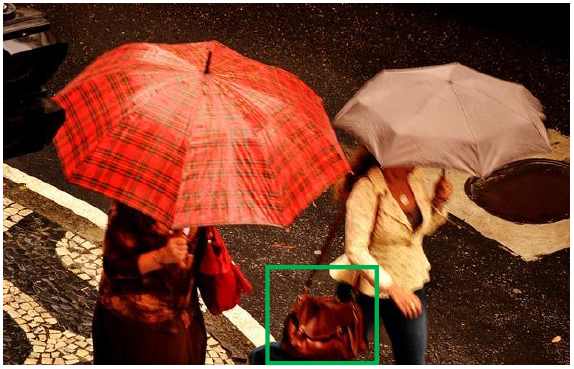}}&Origin: A brown leather handbag.&0.62&0.79&0.78&\green{1.0}\\
&1: A brown \red{metal} handbag.&0.45&0.88&0.56&0.37\\
&2: A brown \red{text} handbag.&0.53&\green{1.0}&0.90&0.50\\
&3: A brown \red{wool} handbag.&0.0&0.0&0.63&0.13\\
&4: A \red{orange} leather handbag.&0.32&0.74&0.34&0.53\\
&5: A brown \red{paper} handbag.&0.41&0.87&\green{1.0}&0.60\\
&6: A \red{light orange} leather handbag.&0.08&0.72&0.08&0.56\\
&7: A brown \red{glass} handbag.&\green{1.0}&0.59&0.75&0.0\\
&8: A \red{dark red} leather handbag.&0.43&0.59&0.15&0.93\\
&9: A \red{dark yellow} leather handbag.&0.31&0.82&0.11&0.73\\
&10: A \red{purple} leather handbag.&0.21&0.91&0.0&0.27\\

\midrule
\multirow{11}{*}{\includegraphics[width=0.25\linewidth]{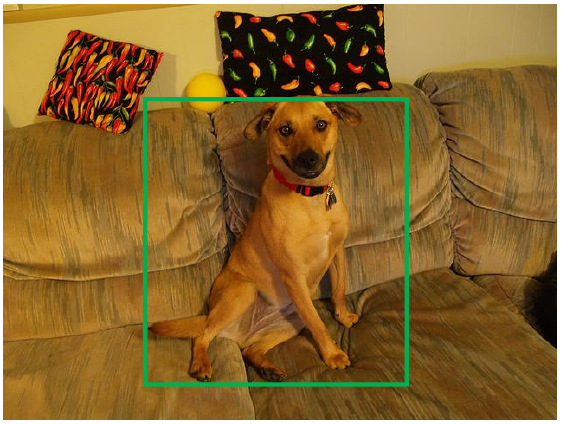}}&Origin: A brown dog with black nose.&0.75&0.0&0.76&\green{1.0}\\
&1: A brown dog with \red{light red} nose.&0.37&0.45&0.64&0.80\\
&2: A brown dog with \red{dark yellow} nose.&0.10&0.39&0.93&0.90\\
&3: A \red{light blue} dog with black nose.&0.40&0.39&0.0&0.0\\
&4: A brown dog with \red{yellow} nose.&0.38&0.40&\green{1.0}&0.86\\
&5: A \red{red} dog with black nose.&\green{1.0}&\green{1.0}&0.59&0.32\\
&6: A brown dog with \red{light orange} nose.&0.16&0.67&0.71&0.85\\
&7: A brown dog with \red{light yellow} nose.&0.0&0.58&0.83&0.83\\
&8: A brown dog with \red{light blue} nose.&0.26&0.44&0.48&0.52\\
&9: A \red{dark green} dog with black nose.&0.88&0.32&0.41&0.13\\
&10: A brown dog with \red{dark purple} nose.&0.41&0.12&0.48&0.73\\

\midrule
\multirow{11}{*}{\includegraphics[width=0.14\linewidth]{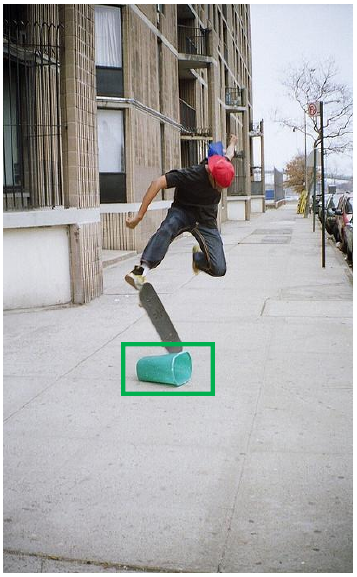}}&Origin: A light blue plastic trash can.&0.89&0.95&0.52&\green{1.0}\\
&1: A light blue \red{stone} trash can.&0.90&\green{1.0}&\green{1.0}&0.77\\
&2: A \red{dark purple} plastic trash can.&0.60&0.70&0.03&0.64\\
&3: A light blue \red{wool} trash can.&0.68&0.57&0.37&0.29\\
&4: A \red{dark green} plastic trash can.&0.92&0.74&0.99&0.65\\
&5: A \red{dark orange} plastic trash can.&0.58&0.35&0.0&0.87\\
&6: A \red{black} plastic trash can.&0.66&0.80&0.65&0.77\\
&7: A \red{purple} plastic trash can.&0.68&0.90&0.36&0.75\\
&8: A light blue \red{crochet} trash can.&0.0&0.0&0.10&0.0\\
&9: A light blue \red{glass} trash can.&\green{1.0}&0.72&0.52&0.64\\
&10: A light \red{orange} plastic trash can.&0.68&0.77&0.02&0.88\\

\midrule
\multirow{11}{*}{\includegraphics[width=0.17\linewidth]{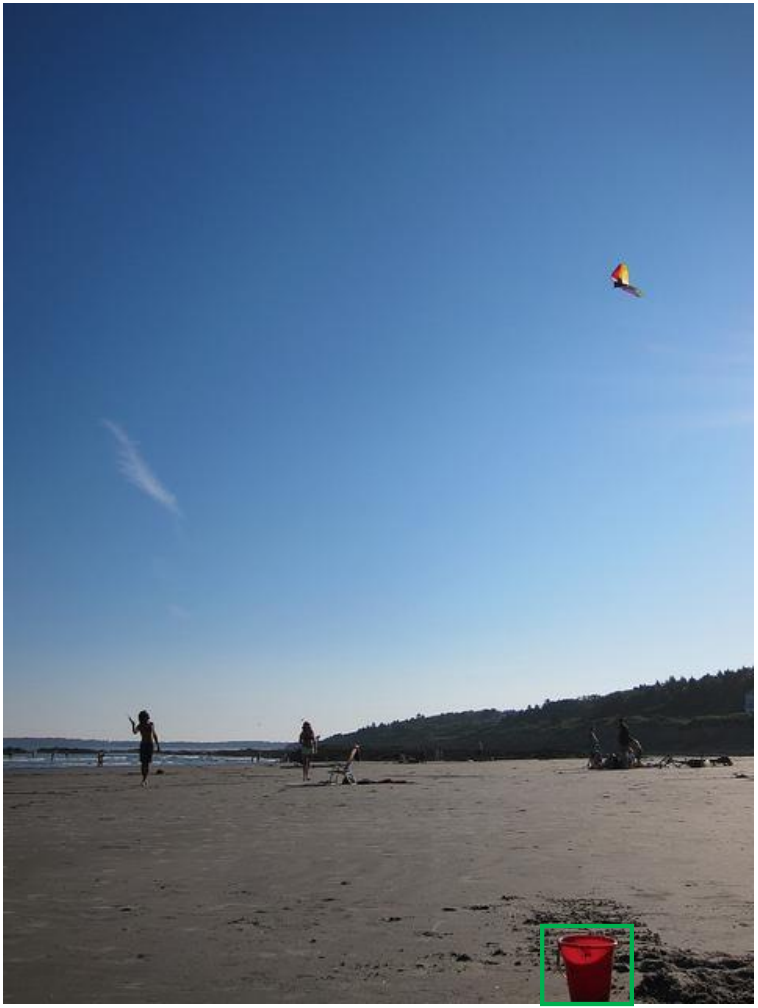}}&Origin: A red plastic bucket.&0.97&0.53&0.70&\green{1.0}\\
&1: A \red{green} plastic bucket.&0.94&0.77&0.46&0.68\\
&2: A red \red{metal} bucket.&0.83&0.29&0.54&0.61\\
&3: A red \red{crochet} bucket.&0.0&0.10&0.10&0.39\\
&4: A red \red{ceramic} bucket.&0.58&0.06&0.15&0.58\\
&5: A red \red{fabric} bucket.&0.55&0.05&0.0&0.43\\
&6: A red \red{stone} bucket.&\green{1.0}&0.10&0.84&0.48\\
&7: A red \red{rattan} bucket.&0.47&0.51&0.58&0.0\\
&8: A red \red{wool} bucket.&0.39&0.0&0.14&0.31\\
&9: A \red{yellow} plastic bucket.&0.77&\green{1.0}&\green{1.0}&0.67\\
&10: A \red{light green} plastic bucket.&0.70&0.87&0.45&0.28\\

\bottomrule
\end{tabular}
}
\end{small}
\end{center}
\vskip -0.2in
\end{table*}

\subsection{Comparison of Different Methods on Fine-Grained Benchmark}
\label{VC2}

\TODO{As shown in Table \ref{fgvc}}, we select several samples from the test set of FG-OVD~\cite{bianchi2024devil} and visualize the comparison results of different methods. We employ the testing strategy detailed in Section \ref{exp_region} and match the text with localized dense feature. The similarity scores computed between regions and texts are normalized, where the sentence with the lowest similarity is assigned a score of 0.0, and the sentence with the highest similarity is assigned a score of 1.0. 
FG-CLIP demonstrates strong capability in identifying these extremely difficult samples, whereas other methods struggle to achieve comparable performance.

\subsection{Performance Comparison on Identical Datasets}

\begin{table}[htbp]
    \centering
    
    \caption{Comparisons of different methods on the same dataset.}
    \vskip 0.1in
    \scalebox{0.9}{
    \begin{tabular}{ccccc}
        \hline
        Method & Data Source & COCO-Box-Top-1 & COCO-Retrieval-I2T & COCO-Retrieval-T2I \\
        \hline
        FineCLIP & FineCLIP (CC2.5M) & 50.7 & 54.4 & 40.2 \\
        FineCLIP & FG-CLIP (12M) & 53.5 & 59.6 & 46.2 \\
        FG-CLIP (Ours) & FG-CLIP (12M) & 56.1 & 65.9 & 47.1 \\
        \hline
    \end{tabular}
    }
    \label{tab:coco_comparison}
\end{table}

\TODO{To evaluate the effectiveness of our proposed FG-CLIP method, we conduct experiments on the same dataset to ensure a fair comparison. Specifically, we compare FineCLIP and FG-CLIP using the 12M dataset due to time constraints, instead of the larger 1.6B+12M setup. From Table \ref{tab:coco_comparison}, the substantial improvements (Row 1 -> Row 2 and Row 2 -> Row 3) highlight that both our proposed dataset and model architecture are significant for FG-CLIP.}

\subsection{Performance on General Multimodal Benchmarks}
\label{addition:mllm}

\begin{table}[htbp]
\caption{Comparisons on General Multimodal
Benchmarks.}
\vskip 0.1in
\begin{center}
\scalebox{0.9}{
\begin{tabular}{lccccccccc}
\toprule
\multirow{2}{*}{Method} & \multirow{2}{*}{GQA} & \multirow{2}{*}{POPE} &\multicolumn{3}{c}{RefCOCO} &\multicolumn{2}{c}{MMBench-EN} &\multicolumn{2}{c}{MMBench-CN}\\
&&&val&testA&testB&dev&test&dev&test \\
\toprule  
LLaVA-v1.5+CLIP&61.9&85.9&76.2&83.4&67.9& 65.1 & 66.5 & 58.2 & 58.4\\
&{\textit{+1.0}}&{\textit{+0.9}}&{\textit{+5.2}}&{\textit{+3.1}}&{\textit{+7.0}}&{\textit{+1.5}}&{\textit{+0.2}}&{\textit{+0.6}}&{\textit{+0.9}}\\
LLaVA-v1.5+FG-CLIP&62.9&86.8&81.4&86.5&74.9& 66.6 & 66.7 & 58.8 & 59.3\\

\bottomrule
\end{tabular}
}
\end{center}
\label{tab:addtion-mllm}
\end{table}

\TODO{In addition to GQA, POPE, and RefCOCO, we conduct experiments on other general multimodal benchmarks~\cite{liu2024mmbench}. The experimental results in Table~\ref{tab:addtion-mllm} show that LLaVA with FG-CLIP achieves better performance.}

\end{document}